\documentclass{article}

\usepackage{microtype}
\usepackage{graphicx}
\usepackage{subcaption}
\usepackage{booktabs} 

\usepackage{hyperref}

\usepackage[preprint]{icml2026}

\usepackage{amsmath}
\usepackage{amssymb}
\usepackage{mathtools}
\usepackage{amsthm}

\usepackage{comment}
\usepackage{graphicx}
\usepackage{multirow}
\usepackage{colortbl}

\usepackage{siunitx}
\usepackage{dcolumn}
\usepackage{tcolorbox}
\usepackage{listings}
\usepackage{xspace}

\usepackage[capitalize,noabbrev]{cleveref}

\theoremstyle{plain}

\theoremstyle{definition}

\theoremstyle{remark}

\usepackage[textsize=tiny]{todonotes}

\icmltitlerunning{MM-THEBench: Do Reasoning MLLMs Think Reasonably?}

\newcommand{\bench}{\textsc{MM-THEBench}\xspace}

\begin{document}

\twocolumn[
  \icmltitle{MM-THEBench: Do Reasoning MLLMs Think Reasonably?}



  \icmlsetsymbol{equal}{*}

  \begin{icmlauthorlist}
    \icmlauthor{Zhidian Huang}{equal,thu}
    \icmlauthor{Zijun Yao}{equal,thu}
    \icmlauthor{Ji Qi}{thu}
    \icmlauthor{Shangqing Tu}{thu}
    \icmlauthor{Junxian Ma}{thu}
    \icmlauthor{Jinxin Liu}{thu} \\
    \icmlauthor{Weichuan Liu}{sms}
    \icmlauthor{Xiaoyin Che}{sms}
    \icmlauthor{Lei Hou}{thu}
    \icmlauthor{Juanzi Li}{thu}
  \end{icmlauthorlist}

  \icmlaffiliation{thu}{DCST, BNRist; KIRC, Institute for Artificial Intelligence, Tsinghua University, China}
  \icmlaffiliation{sms}{Siemens AG, China}


  \icmlkeywords{Machine Learning, ICML}

  \vskip 0.3in
]



\printAffiliationsAndNotice{\icmlEqualContribution}

\begin{abstract}
Recent advances in multimodal large language models (MLLMs) mark a shift from non-thinking models to post-trained reasoning models capable of solving complex problems through thinking.
However, whether such thinking mitigates hallucinations in multimodal perception and reasoning remains unclear.
Self-reflective reasoning enhances robustness but introduces additional hallucinations, 
and subtle perceptual errors still result in incorrect or coincidentally correct answers.
Existing benchmarks primarily focus on models before the emergence of reasoning MLLMs, neglecting the internal thinking process and failing to measure the hallucinations that occur during thinking.
To address these challenges, we introduce \bench, a comprehensive benchmark for assessing hallucinations of intermediate CoTs in reasoning MLLMs.
\bench \quad features a fine-grained taxonomy grounded in cognitive dimensions, diverse data with verified reasoning annotations, and a multi-level automated evaluation framework.
Extensive experiments on mainstream reasoning MLLMs reveal insights into how thinking affects hallucination and reasoning capability in various multimodal tasks.
\end{abstract}
\section{Introduction}
\label{sec:intro}

Reasoning multimodal large language models (reasoning MLLMs)~\cite{openai2024o1,openai2025o3,openai2025GPT-5,Comanici2025Gemini2P,anthropic2025claude4}, developed by incentivizing MLLMs to produce long intermediate chain-of-thoughts (CoTs) before generating their final outputs, have demonstrated remarkable performance on various complex visual reasoning tasks~\cite{wang2024measuring,yue2023mmmu,yu2024mmvetv2,2023arXiv231014566G,jia2025omnispatial,wang2024charxiv,wang2025guiagent,fu2025videomme}.
Although these intermediate reasoning steps provide explainability for model decisions~\citep{baker2025monitoring,korbak2025chain}, there are observations that reasoning models can produce correct answers even with obvious faults in their intermediate CoTs~\citep{arcuschin2025chainofthought,bi2025verify, huang2025answerconsistent,yao2025reasoning}.
Nevertheless, it still lacks a systematic way to monitor these intermediate CoTs.
As CoT monitoring is of vital importance to guide the development of reasoning MLLMs, we aim to address the following research question: \textbf{Can we evaluate the plausibility of intermediate CoTs from reasoning MLLMs?}

Examining the intermediate CoTs of reasoning MLLMs is challenging for two main reasons.
\textbf{(1) Lack of a hallucination taxonomy for intermediate CoTs}.
Existing benchmarks for evaluating reasoning MLLMs focus primarily on the correctness of final outputs.
Incorrect final answers are often roughly attributed to \textit{hallucinations} in intermediate CoTs~\citep{2023arXiv231014566G}, which are defined as contents that are inconsistent with factual knowledge, the given multimodal evidence, or the logical context~\cite{huang2025survey,wang2025comprehensive}.
However, which part of the intermediate CoTs is hallucinated is rarely analyzed.
Most recent studies~\cite{2025arXiv250524238D, jiang2025mme} aim to construct benchmarks that evaluate reasoning MLLMs through stepwise analysis of intermediate CoTs.
However, they do not provide a comprehensive taxonomy of hallucination categories that occur in these CoTs, resulting in a coarse-grained understanding of intermediate reasoning failures.

Moreover, the evaluation of intermediate CoTs is hindered by \textbf{(2) free-form and long-form generation}.
Intermediate CoTs generated by reasoning MLLMs often comprise thousands of tokens and lack a fixed structure.
This makes it difficult to design simple rule-based verifiers~\citep{yang2025r1onevision,ouyang2025spacer,xiao2025advancing,wang2025vl,peng2025skywork} that are commonly used to examine the correctness of the final answer.
Although human inspection is possible for case studies~\citep{yu2024rlhfv,zhang2025mm}, manually annotating the correctness of steps in the intermediate CoTs at scale is impractical.

\begin{figure*}[t]
    \centering
    \includegraphics[width=0.95\linewidth]{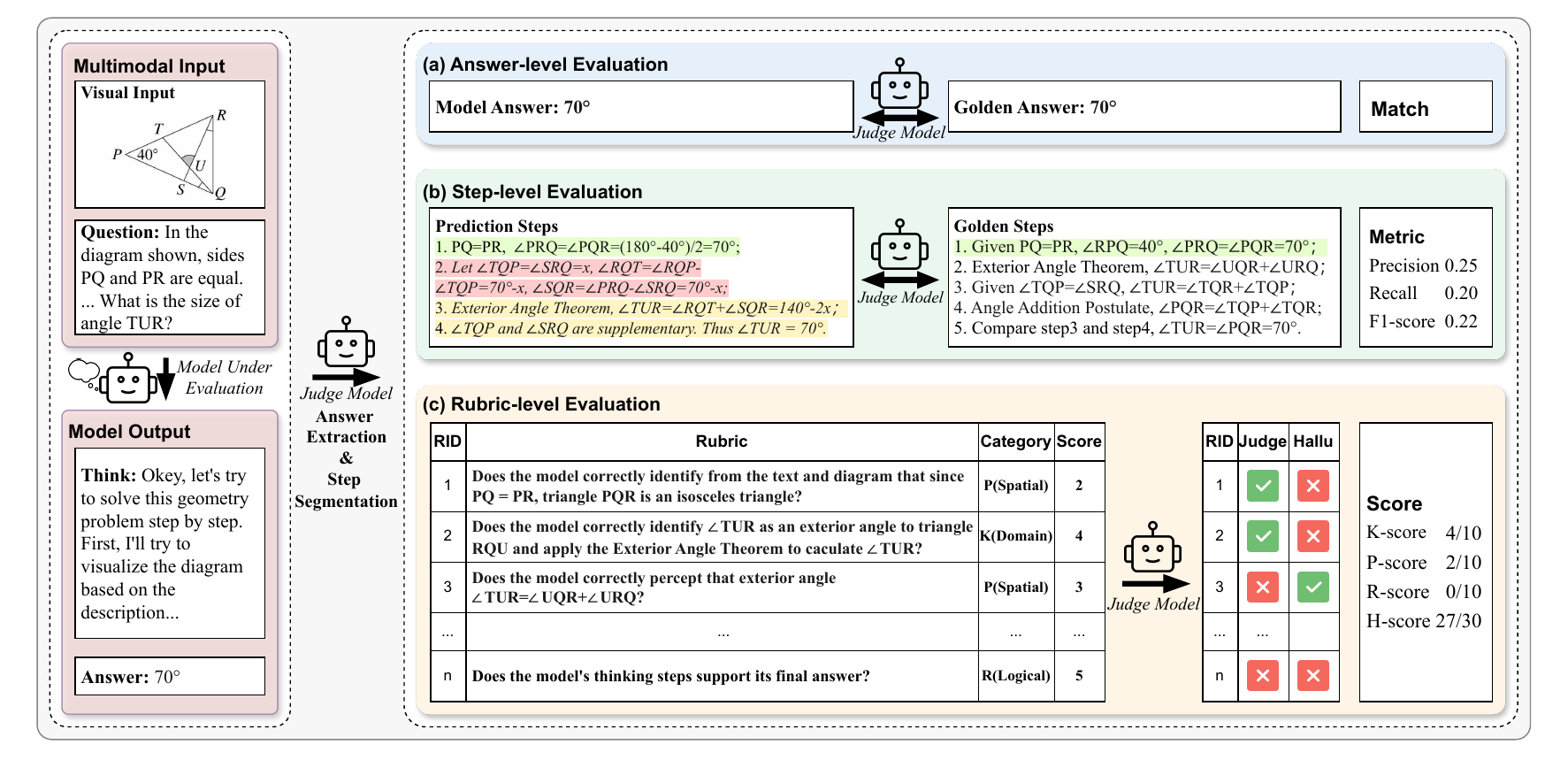}
    \vspace{-0.1in}
    \caption{Overview of the evaluation framework in \bench. 
    In (c) Rubric-level Evaluation, the column ``RID'' denotes the rubric item ID. 
    ``Judge''  indicates whether the judge model considers the item satisfied in the intermediate CoT,
    and ``Hallu'' indicates whether a hallucination occurs for the corresponding rubric item.
    ``K'', ``P'', ``R'', represents the dimensions Knowledge, Perception, and Reasoning.
    ``H-score'' denotes Hallucination-free score (1-hallucination score ratio) .}
    \label{fig:framework}
\end{figure*}

To address these challenges, this paper introduces \bench~(\underline{M}ulti\underline{m}odal \underline{t}hinking \underline{h}allucination \underline{e}valuation \underline{bench}mark), a comprehensive benchmark designed to systematically assess hallucinations in the intermediate CoTs produced by reasoning MLLMs.
\bench features a fine-grained hallucination taxonomy grounded in cognitive dimensions of multimodal reasoning, along with a multi-level automated evaluation framework.
We construct \bench by annotating verified reasoning steps across diverse multimodal data.
The overall evaluation framework is illustrated in \cref{fig:framework}.

\bench defines a two-layer hallucination taxonomy for intermediate CoTs.
At the top level, it specifies three cognitive dimensions---knowledge, perception, and reasoning---each of which is further subdivided into fine-grained subcategories.
To automatically perform step-wise evaluation on free-form intermediate CoTs, we implement a rubric-based evaluation~\cite{peng2025verif,huang2025reinforcement,liu2025openrubrics,sharma2025researchrubrics} protocol that quantifies thinking quality and hallucination occurrence across these cognitive dimensions.
The constructed \bench contains $1,340$ diverse multimodal questions spanning images and videos.
For each question, \bench provides annotations of the necessary atomic reasoning steps along with verified chains of thought.

We comprehensively evaluate $14$ latest reasoning MLLMs on \bench, including GPT-5~\cite{openai2025GPT-5} and OpenAI-o3~\cite{openai2025o3}.
Our evaluation results show that the correctness of intermediate CoTs from these state-of-the-art reasoning MLLMs lags significantly behind the accuracy evaluated against their final answers, suggesting that these intermediate CoTs fail to provide a faithful explanation of the true decision reasons.
We further analyze and show that the correctness of the final answer is far more sensitive to reasoning hallucinations than to perception hallucinations.
Perception hallucinations happen most frequently among the three cognitive dimensions.
However, perception hallucinations rarely lead to wrong answers, whereas reasoning or mixed hallucinations show a much stronger association with incorrect outcomes across all models.
Under the guidance of our proposed hallucination taxonomy, we find that spatial-related hallucinations dominate across the hallucination subcategories in both perception and reasoning dimensions.
These experimental results call for a detailed monitoring of the intermediate CoTs in reasoning MLLMs in future research.

Our contributions are summarized as follows:
(1) We design and construct \bench, a comprehensive benchmark with a fine-grained taxonomy to evaluate thinking hallucination in multimodal reasoning; 
(2) We develop a highly automated, multi-level evaluation framework to quantify hallucination categories, facilitating future research and model assessment; 
(3) We evaluate the latest mainstream reasoning MLLMs on \bench, revealing similar hallucination patterns across different models.

\section{The Proposed \bench}
\label{sec:bench}
We introduce \bench, a benchmark designed to assess the thinking capabilities and hallucinations of reasoning MLLMs.
We first introduce our design philosophy, including the hallucination taxonomy.
Then, we collect multimodal data and manually annotate necessary atomic steps in intermediate CoTs with strict quality control.
Finally, we perform statistical analysis for \bench.

\subsection{Benchmark Design}
\label{subsec:benchmark_design}
\bench is designed to evaluate step-level hallucinations in intermediate CoTs produced by reasoning MLLMs.
To this end, \bench is presented as a multimodal question answering task.
Each question is associated with identified non-omittable atomic reasoning steps.
For each step, we further annotate rubrics to guide LLM judges in comparing intermediate CoTs to the annotated reference reasoning step.
Rather than expanding data volume, \bench focuses on transforming existing high-quality benchmarks into a process-aware evaluation resource through fine-grained annotation.

We identify three core capabilities that a reasoning MLLMs should possess: knowledge, perception, and reasoning~\cite{yue2023mmmu,zhao2024benchmarking,wu2025combating}.
These dimensions not only capture the essential aspects of multimodal cognitive processing but also align with the primary categories of hallucination that may arise in the intermediate CoTs of reasoning MLLMs.
Specifically, each top-level dimension is divided into subcategories:
$\bullet$ \textbf{Knowledge} is further divided into \textit{world knowledge}, \textit{commonsense}, and \textit{domain knowledge}.
$\bullet$ \textbf{Perception} is divided into \textit{recognition}, \textit{OCR}, \textit{spatial}, \textit{count}, \textit{audio}, \textit{grounding}, \textit{temporal}.
$\bullet$ \textbf{Reasoning} is divided into \textit{deductive}, \textit{inductive}, \textit{spatial}, \textit{arithmetic}, \textit{causal}, \textit{decision}, \textit{instructional}.
This hierarchical design enables \bench to comprehensively assess both the capabilities of reasoning MLLMs and the prevalence of hallucinations across cognitive dimensions.
Detailed definitions are provided in Appendix \ref{app:taxonomy}.

\subsection{Data Collection}
\label{subsec:data_collection}
All samples in \bench are derived from existing high-quality datasets to enable consistent evaluation across tasks and domains.
We select representative datasets with complementary focuses to ensure balanced coverage of diverse task types, including: 
(1) \textbf{MathVision}~\cite{wang2024measuring} for symbolic perception and mathematical reasoning;
(2) \textbf{MM-vet-v2}~\cite{yu2024mmvetv2} for multi-image integrative reasoning;
(3) \textbf{MMMU-pro}~\cite{yue2023mmmu} for multi-domain knowledge reasoning;
(4) \textbf{HallusionBench}~\cite{2023arXiv231014566G} for hallucination and counterfactual perception;
(5) \textbf{Omni-Spatial}~\cite{chen2024spatialvlm} for spatial visual reasoning;
(6) \textbf{CharXiv}~\cite{wang2024charxiv} for scientific chart understanding;
(7) \textbf{GUI-Agent}~\cite{wang2025guiagent} for graphical user interface (GUI) grounding;
and (8) \textbf{Video-MME}~\cite{fu2025videomme} for temporal reasoning from videos.

Prior to integration, all samples undergo filtering to ensure diversity and balance across modalities (single-image, multi-image, and video) and to remove low-quality or inaccessible items. 
Details are provided in Appendix \ref{app:construction}.

\subsection{Data Annotation and Quality Control}
\label{subsec:annotation}
\textbf{Step Annotation.}
Since the raw data lacks grounded reasoning steps, we systematically annotate the necessary thinking processes.
We leverage the state-of-the-art reasoning MLLM, Gemini-2.5-pro~\cite{Comanici2025Gemini2P}, 
to automatically generate step-by-step reasoning paths for each question.
These steps capture the essential perceptual observations and logical transitions underlying the correct final answer.
Each reasoning step is further labeled with its corresponding dimension: Knowledge, Perception, Reasoning, or their combination.
This categorization facilitates subsequent analysis of hallucinations across these dimensions.

\textbf{Human Verification.}
Human annotators review and revise the generated reasoning steps to ensure correctness, completeness, and necessity. 
Ambiguous questions and errors in the original dataset are corrected during this process.
This hybrid annotation pipeline combines the scalability of automated generation with the reliability of human supervision.

\textbf{Rubric Annotation.}
Based on the verified reasoning chains, we prompt the annotation model to construct a fine-grained rubric for each question to evaluate thinking quality.
Each rubric item corresponds to a key atomic reasoning step, including: 
(1) a binary tag indicating whether the rubric is satisfied in the intermediate CoTs from reasoning MLLMs; 
(2) a score reflecting the required capability or difficulty level to satisfy this item, and
(3) a semantic label from the three cognitive dimensions, along with a specific subcategory.
This structured design enables interpretable and quantitative assessment of reasoning quality and hallucination behaviors across different dimensions.

\textbf{Quality Control.}
We adopt a two-stage quality control protocol.
In the data annotation stage, we sample $10\%$ annotations for each annotator.
If more than $30\%$ of the instances fail our quality check, we discard all data from the examined annotator and relabel them.
In the final stage, we resample $10\%$ of the data and ensure that fewer than $30\%$ are labeled as disqualified.

\subsection{Data Statistics and Analysis}
\label{subsec:data_statistics}

\begin{figure}[t]
  \centering
  \begin{subfigure}[b]{0.98\linewidth}
    \centering
    \includegraphics[width=\linewidth]{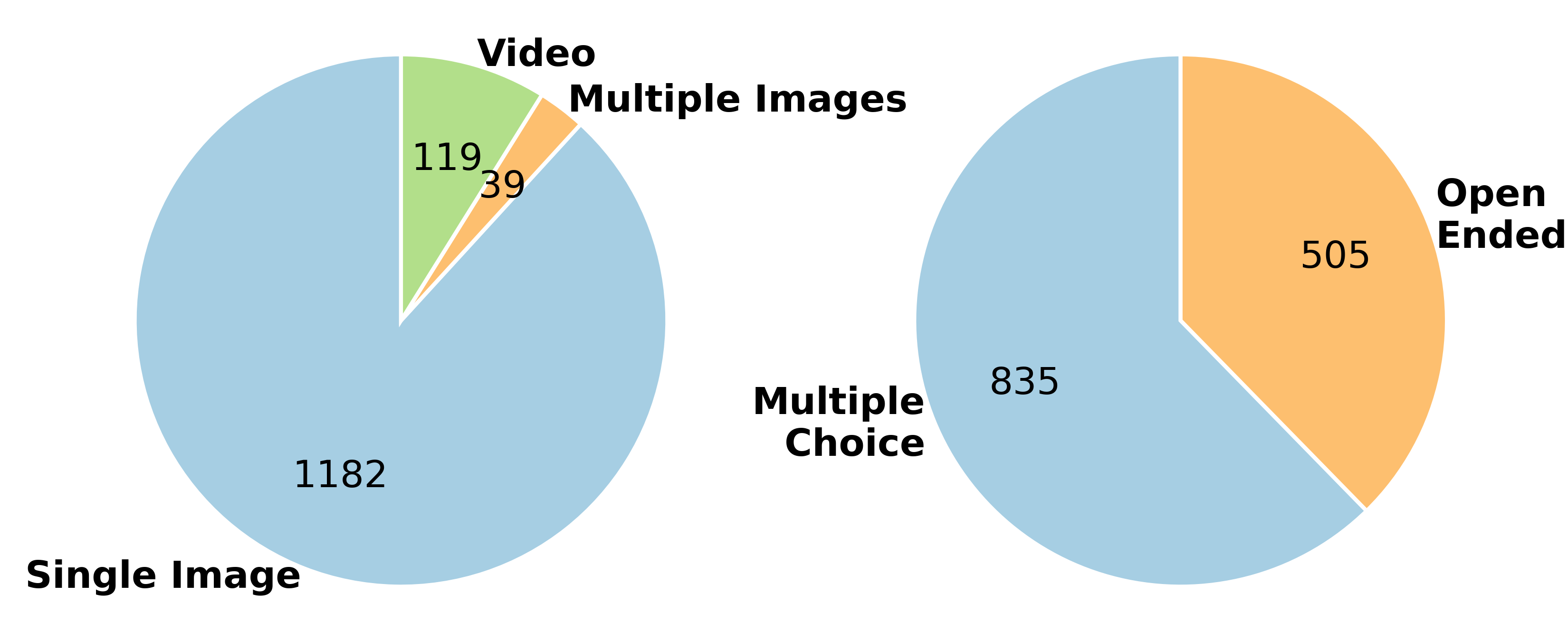}
  \end{subfigure}
  \caption{Data composition in \bench. \textbf{Left}: Modality composition. \textbf{Right}: Question format composition.}
  \label{fig:dataset_composition}
\end{figure}

\textbf{Data Composition.}
We analyze \bench in terms of modality and question format.
The final dataset contains $1,182$ single-image samples, $39$ multi-image samples, and $119$ video samples, covering diverse multimodal contexts, as illustrated in \cref{fig:dataset_composition}.
Regarding question format, \bench consists of $835$ multiple-choice and $505$ open-ended questions, allowing both structured and generative evaluations of reasoning performance.

\begin{table}[t]
\centering
\caption{Step-level statistics across the three cognitive dimensions. (A single step may involve multiple dimensions, so the sum of 3 percentages exceeds 100\%.)}
\label{tab:step_level_stats}
\begin{tabular}{ccc}
\toprule
\textbf{Type} & \textbf{Step Ratio(\%)} & \textbf{Avg. Steps / Q.} \\
\midrule
Knowledge  & $15.30$ & $0.63$  \\
Perception & $49.71$ & $2.08$  \\
Reasoning  & $60.31$ & $2.52$  \\
\midrule
All Steps  & $-$      & $4.17$  \\
\bottomrule
\end{tabular}
\end{table}
\textbf{Step Analysis.}
To further characterize the annotated reasoning processes, we perform a step-level analysis that quantifies the proportion of thinking steps associated with different dimensions, as shown in \cref{tab:step_level_stats}.
Note that a single step may involve multiple dimensions. 
For example, identifying an object in a given image as the \textit{Eiffel Tower} requires both visual perception and world knowledge.
Consequently, the aggregated proportions across dimensions may exceed 100\%.
In addition, we report the average total number of thinking steps per question, as well as the average number of steps associated with each dimension, 
providing a more complete characterization of the step-level annotations.

\section{The Proposed Evaluation Framework}
\label{subsec:evaluation_framework}

We propose an LLM-as-a-Judge~\cite{gu2025surveyllmasajudge, li2025generationjudgmentopportunitieschallenges,chen2024humans} based framework to provide rigorous evaluation for free-form and long-form intermediate CoTs in \bench.

\subsection{Judge Model}
To ensure consistent and scalable evaluation of thinking quality and hallucination, we adopt LLM-as-a-judge in the evaluation pipeline of \bench.
Specifically, we employ Qwen-3-32B~\cite{qwen3technicalreport}, which exhibits strong instruction-following and judgment capabilities.
Since our benchmark provides explicit thinking steps for each question, the judging process operates entirely in the textual domain without requiring visual input.
Beyond its evaluation quality, Qwen-3-32B is chosen for its practicality:
It can be deployed on laboratory-scale compute resources, ensuring an economically feasible and fully reproducible evaluation pipeline without reliance on external APIs.
Within our evaluation framework, the judge model undertakes four complementary responsibilities:
(1) answer extraction and verification; 
(2) segmentation of reasoning steps;
(3) step matching; and
(4) rubric-based evaluation.
This design enables the judge model to function as a unified evaluator across the entire assessment pipeline, 
ensuring consistent interpretation of reasoning and stable quantification of hallucination across diverse multimodal tasks.

\begin{table*}[t]
\centering
\small
\caption{Overall performance of reasoning MLLMs on image-based tasks across three evaluation levels. 
Columns Acc, Precision, Recall, and F1 represent percentage values (\%).
Columns K, P, and R represent the normalized scores (0-100) in the dimensions of Knowledge, Perception, and Reasoning, and H denotes the H-score.
Models are divided into two groups: models with accessible intermediate CoTs (normal background) and models with invisible intermediate CoTs (gray-shaded rows).}
\label{tab:overall_performance}
\scalebox{0.95}{
\setlength{\tabcolsep}{7.5pt}
\begin{tabular}{lcccccccc}
\toprule
\multirow{2}{*}{\makebox[100pt][c]{\textbf{Model}}} & \multicolumn{1}{c}{\textbf{Answer}} & \multicolumn{3}{c}{\textbf{Step}} & \multicolumn{4}{c}{\textbf{Rubric}} \\
                     \cmidrule(r){2-2} \cmidrule(lr){3-5} \cmidrule(lr){6-9}
                     {} & \textbf{Acc} & \textbf{Precision} & \textbf{Recall} & \textbf{F1} & \textbf{K} & \textbf{P} & \textbf{R} & \textbf{H}\\
\midrule     
                    MiniCPM-4.5V          & $50.76$ & $17.83$ & $42.54$ & $24.11$ & $61.08$ & $52.86$ & $52.73$ & $90.02$ \\
                    GLM-4.1V-Thinking    & $53.88$ & $20.38$ & $40.82$ & $26.20$ & $55.84$ & $56.78$ & $51.04$ & $92.25$ \\
                    Qwen3-VL-8B-Thinking  & $60.03$ & $20.13$ & $51.58$ & $27.42$ & $66.64$ & $61.67$ & $67.28$ & $91.61$ \\
                    InternVL3.5-8B        & $54.24$ & $17.43$ & $46.30$ & $23.89$ & $69.47$ & $57.73$ & $61.97$ & $91.22$ \\
                    GLM-4.5V             & $63.07$ & $23.15$ & $47.05$ & $29.86$ & $64.50$ & $57.17$ & $63.80$ & $92.76 $\\
                    Qwen3-VL-235B-A22B-Thinking & $\mathbf{70.62}$ & $23.38$ & $\mathbf{58.34}$ & $31.83$ & $78.66$ & $69.30$ & $\mathbf{75.00}$ & $\mathbf{94.11}$ \\
                    InternVL3.5-241B-A28B & $62.47$ & $21.37$ & $54.21$ & $29.15$ & $\mathbf{78.73}$ & $69.66$ & $70.07$ & $93.98$ \\
                    Doubao-seed-1.6       & $63.07$ & $24.30$ & $51.69$ & $31.96$ & $76.12$ & $63.28$ & $70.66$ & $92.71$\\
                    Claude-Opus-4.1       & $61.07$ & $\mathbf{25.56}$ & $53.61$ & $\mathbf{33.24}$& $74.52$ & $68.07$ &  $66.67$ & $93.38$\\
                    
                    Claude-Sonnet-4.5     & $61.19$ & $21.87$ & $56.23$ & $30.28$ & $77.41$ & $\mathbf{70.10}$ & $69.84$ & $92.42$\\
\midrule 
\rowcolor{gray!20}
                    o3                    & $65.55$ & $37.50$ & $59.57$ & $44.77$ & $79.09$ & $71.84$ & $76.42$ & $95.01$ \\
\rowcolor{gray!20}
                    GPT-5                 & $67.87$ & $39.56$ & $59.19$ & $46.17$ & $81.70$ & $68.36$ & $77.03$ & $93.97$ \\      
\rowcolor{gray!20}
                    Gemini-2.5-Flash      & $65.35$ & $33.15$ & $62.76$ & $42.08$ & $86.16$ & $75.94$ & $80.03$ & $94.79$ \\
\rowcolor{gray!20}
                    Gemini-2.5-Pro        & $67.31$ & $39.88$ & $70.21$ & $49.41$ & $90.83$ & $80.52$ & $83.60$ & $95.68$ \\
\bottomrule
\end{tabular}
}
\end{table*}

\subsection{Evaluation Pipeline}
To ensure fair and consistent evaluation across different MLLMs, we design a unified prompting protocol that elicits both the final answer and the underlying CoT.
For models supporting explicit reasoning outputs (\textit{e.g.,} API models with ``thinking'' outputs or open-source models), we directly extract their internal CoTs and final responses.
For closed-source API models where intermediate reasoning is inaccessible, we approximate their thinking process by parsing the reasoning steps summarized in their final responses, 
model-specific configurations are detailed in \cref{sec:experiments}.

Building on this unified setup, the evaluation pipeline of \bench proceeds through three stages:

\textbf{(1) Answer-level Evaluation}:
The judge model verifies the predicted final answer, ensuring robust answer evaluation across varied output formats, shown in \cref{fig:framework} (a).
\textbf{(2) Step-level Evaluation}: 
The reasoning text is segmented into atomic steps and aligned with the annotated reasoning chain, shown in \cref{fig:framework} (b).
This stage assesses the structural fidelity of the model's reasoning process and identifies incorrect steps.
\textbf{(3) Rubric-level Evaluation}: 
Using rubrics aligned with the three dimensions, the judge model assigns fine-grained scores to each CoT, shown in \cref{fig:framework} (c).

Together, these three levels form a unified and interpretable framework for evaluating reasoning MLLMs, enabling disentangled analysis of answer accuracy, step consistency, and reasoning hallucination.
More details are shown in \ref{app:framework}.

\subsection{Evaluation Metrics}
We define evaluation metrics at three levels corresponding to the assessment stages of \bench.
At the \textbf{answer level}, correctness is computed based on task format:
Multiple-choice accuracy is measured by the proportion of correct predictions.
open-ended responses are judged as MATCH, PARTIAL, or MISMATCH, and accuracy is
 \begin{equation}
    \text{ACC} = \frac{N_\text{MATCH}+0.5 \times N_\text{PARTIAL} }{N_\text{total}}
\end{equation}
For grounding tasks, correctness is measured using the Intersection over Union (IoU) metric with a threshold of $0.5$.

At the \textbf{step level}, we adopt Precision, Recall, and F1-score to measure alignment between model-generated reasoning steps and annotated steps.
Precision reflects the proportion of correct and relevant steps,
while recall captures how many necessary steps are reproduced.
F1-score summarizes the structural fidelity of the reasoning process.

At the \textbf{rubric level}, models' capabilities and hallucinations are quantified using rubric-based scores across the three dimensions.
Scores are normalized per dimension,
and their subcategories support fine-grained analysis of error patterns and hallucination types.
We further define the \textbf{H-score} as the hallucination-free score, computed as $1-$(hallucination score ratio),
quantifying the absence of hallucinations.

Finally, we also measure \textbf{thinking length} using the annotated reasoning chain as the necessary length and treat each model's generated token length as the absolute length. 
Then the relative thinking length is computed as the ratio of absolute length to necessary length.

\section{Experiments}
\label{sec:experiments}

We conduct experiments to evaluate state-of-the-art reasoning MLLMs on \bench and analyze the quality of their intermediate CoTs.

\subsection{Experimental Setup}
\label{subsec:expr_setup}
\textbf{Models.} 
We categorize the evaluated reasoning MLLMs into two groups:
(1) \textbf{Models with accessible intermediate CoTs.}
This group includes open-source models and some closed-source models exposing internal thinking.
We evaluate MiniCPM-4.5V~\cite{yao2024minicpm}, 
GLM-4.1V-thinking, GLM-4.5V~\cite{vteam2025glm45v},
Qwen3-VL-8B-Thinking, Qwen3-VL-235B-A22B-Thinking~\cite{qwen3technicalreport, Qwen2.5-VL},
InternVL3.5-8B, InternVL3.5-241B-A28B~\cite{wang2025internvl3_5},
Doubao-seed-1.6~\cite{guo2025seed15vl}, 
Claude-Opus-4.1, and Claude-Sonnet-4.5~\cite{anthropic2025claude4}. 
For these models, the true intermediate CoTs can be obtained directly through deployment or official APIs, accessing genuine perception and reasoning chains.
(2) \textbf{Models with invisible intermediate CoTs.}
These models enforce stricter confidentiality, and their internal thinking processes are not visible to users.
To approximate their reasoning process, we employ a CoT prompting strategy, instructing the models to produce step-by-step reasoning in their final outputs to reconstruct their original intermediate CoTs.
We evaluate GPT-5~\cite{openai2025GPT-5}, o3~\cite{openai2025o3}, Gemini-2.5-Pro, and Gemini-2.5-Flash~\cite{Comanici2025Gemini2P}.

\textbf{Implementation Details.}
To ensure a fair comparison and faithfully reflect each model's inherent capabilities, we use the official APIs whenever available, including for open-source models.
All models are evaluated using their officially recommended default parameters (\textit{e.g.,} temperature). 
For models that provide an explicit thinking mode or an equivalent reasoning switch, we enable it during evaluation to fully expose their reasoning behaviors and hallucination tendencies under standard conditions.

\textbf{Meta-evaluation of Judge Model.}
To validate the reliability of the judge model, we conduct a human meta-evaluation on a randomly sampled subset of 300 evaluation instances.
In each instance, human experts follow the same judgment procedure using identical evaluation principles.
We then compute the same metrics based on human judgments and compare them with the corresponding judge model estimates.
Overall, the judge model shows strong agreement with human evaluations across all metrics.
Notably, at the answer level, the accuracy deviation is below 1\%, indicating high certainty in evaluating final answer correctness, while deviations at finer-grained levels remain within a small and consistent range.

\subsection{Overall Performance on Image-Based Data}
\label{subsec:overall_performance}

\cref{tab:overall_performance} presents the overall results of all evaluated reasoning MLLMs on image-based data across three levels.
Video-based evaluations are excluded in this subsection, as several models lack support for temporal visual inputs or corresponding API interfaces.
A separate analysis of video data is provided later.
Unless otherwise specified, all analyses in this section are conducted on the full image-based dataset.

\textbf{Answer-level Analysis.}
Qwen3-VL-235B-A22B-Thinking achieves the highest accuracy ($70.62\%$), followed by GPT-5.
Among 8B-scale models, Qwen3-VL-8B-Thinking attains $60.03\%$ accuracy, outperforming InternVL3.5-8B, MiniCPM-4.5V, and GLM-4.1V-Thinking, showing that smaller models can achieve competitive reasoning performance.
Larger models, such as GLM-4.5V and InternVL3.5-241B-A28B, show further improvements, highlighting the effects of model scale.

\textbf{Step-level Analysis.}
Step-level metrics assess the alignment between the model-generated reasoning steps and the annotated ground-truth steps.
For models with explicit thinking, precision ranges from $17.43\%$ to $25.56\%$, reflecting only a moderate proportion of the thinking steps necessary and correct, with some redundancy or verbosity.
Recall ranges from $40.82\%$ to $58.34\%$, indicating good coverage of essential steps.
In contrast, models evaluated via CoT achieve higher precision, demonstrating more focused and high-utility step generation in the final response.
However, the recall advantage is less pronounced compared to models with explicit thinking.
Overall, CoT yields more efficient reasoning chains without losing necessary step coverage.

\textbf{Rubric-level Analysis.}
With rubric-level metrics, we decompose performance into the three dimensions of Knowledge, Perception, and Reasoning.
The differences between explicit-thinking and CoT models become clearer.
Models with explicit thinking exhibit balanced scores across three dimensions, achieving moderate reasoning.
While reasoning MLLMs excel in knowledge utilization and reasoning, they fall slightly short in perception, highlighting challenges in fine-grained visual understanding.
In addition, the hallucination-free score (Column H) shows that most thinking steps in explicit-thinking models are free from hallucination, with values generally above $90$,
while the reasoning chains provided by CoT models achieve slightly higher H-scores. 
Note that Gemini-2.5-Pro serves as the annotation model in \bench. 
Therefore, its results are primarily used to validate the annotation quality rather than as evidence of superior reasoning capabilities.

\subsection{Overall Performance on Video-Based Data}
\label{subsec:video_performance}
\begin{table}[t]
\centering
\caption{Performance of reasoning MLLMs on video data. 
The meanings of each column are the same as those in \cref{tab:overall_performance}}
\label{tab:video_performance}
\resizebox{\linewidth}{!}{
\begin{tabular}{lccccc}
\toprule
\textbf{Model} & \textbf{Acc} & \textbf{F1} & \textbf{P} & \textbf{R} & \textbf{H}\\
\midrule
GLM-4.1V-Thinking & $40.80$ & $15.82$ & $30.53$ & $50.19$ & $87.73$ \\
GLM-4.5V          & $51.20$ & $\mathbf{36.65}$ & $\mathbf{51.11}$ & $62.49$ & $89.42$ \\
Doubao-seed-1.6    & $\mathbf{61.60}$ & $24.14$ & $49.13$ & $\mathbf{64.91}$ & $\mathbf{90.84}$ \\
\midrule 
\rowcolor{gray!20}
Gemini-2.5-Flash   & $68.80$ & $43.90$ & $70.02$ & $76.62$ & $95.16$ \\
\rowcolor{gray!20}
Gemini-2.5-Pro     & $81.60$ & $53.56$ & $85.50$ & $91.71$ & $96.52$ \\
\bottomrule
\end{tabular}
}
\end{table}
The results of the evaluation on video-based data are presented in \cref{tab:video_performance}.
We find that: (1) Considering the accuracy of the two models in the CoT group, the Gemini-2.5 series maintains a significant advantage over other models.
(2) The performance comparison between the two GLM models with different parameter scales reveals a substantial improvement in perception and reasoning capabilities on video data as the size increases.
(3) Furthermore, although GLM-4.5V has higher thinking quality in terms of F1-score compared with Doubao-seed-1.6, 
the latter achieves noticeably higher accuracy, which may be attributed to its stronger video reasoning and hallucination resistance abilities.

\subsection{Hallucination Analysis}
\label{subsec:hallu_analysis}

\begin{figure}[t]
  \centering
  \begin{subfigure}[b]{0.9\linewidth}
    \centering
    \includegraphics[width=\linewidth]{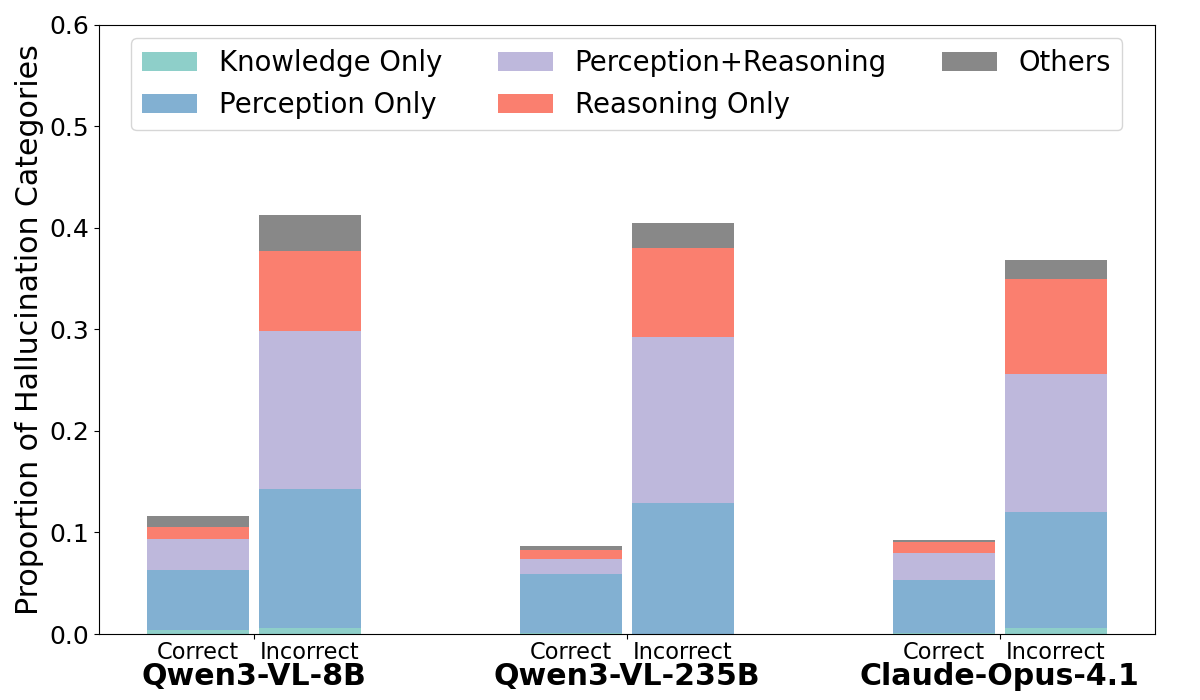}
  \end{subfigure}
  \vspace{-0.1in}
  \caption{
  Hallucination category distribution by model and answer correctness. 
  Qwen3-VL-8B denotes Qwen3-VL-8B-Thinking, and Qwen3-VL-235B denotes Qwen3-VL-235B-A22B-Thinking.}
\label{fig:acc_hallu}
\end{figure}

\textbf{Impact of Hallucinations on Answer Accuracy.}
It is evident that the presence of hallucinations does not always result in incorrect answers.
We show a case in \cref{fig:framework}, even when a model exhibits notable hallucinations in its intermediate CoTs, it may still produce the correct answer.

To further quantify the distribution of different hallucination types with respect to answer correctness, we plot the proportion of hallucination types for each reasoning MLLM in a stacked bar chart, as shown in \cref{fig:acc_hallu}.
Each model is represented by two adjacent bars, one for the correct and one for the incorrect final answers.
The partial-correct cases are omitted in this figure, as they account for less than 1\% of the data.
Each bar corresponds to the fraction of data points exhibiting hallucinations in the thinking process, 
with segments indicating the distribution of different dimensions.

We draw three key conclusions from the visualization:
(1) A correct final answer does not imply an absence of hallucinations in the model's intermediate CoTs. 
For correct answers, hallucination rates remain low (below $15\%$), much lower than for incorrect answers (above $30\%$);
(2) Compared to hallucinations occurring solely in the Perception or Reasoning dimensions, 
the co-occurrence of both categories within the same data point (purple segment) is more frequent in these models.
In contrast, Knowledge-only hallucinations (green) and other compound hallucinations (gray) are relatively rare;
(3) The ratios of \textbf{Reasoning + Perception} and \textbf{Reasoning Only} hallucinations are lower for correct answers than for incorrect ones,
whereas \textbf{Perception Only} hallucinations show weaker correlation with answer correctness.
This suggests that hallucinations in the reasoning dimension tend to have a greater impact on final answer correctness than hallucinations in the perception dimension.

\begin{figure}[t]
  \centering
  \begin{subfigure}[b]{0.90\linewidth}
    \centering
    \includegraphics[width=\linewidth]{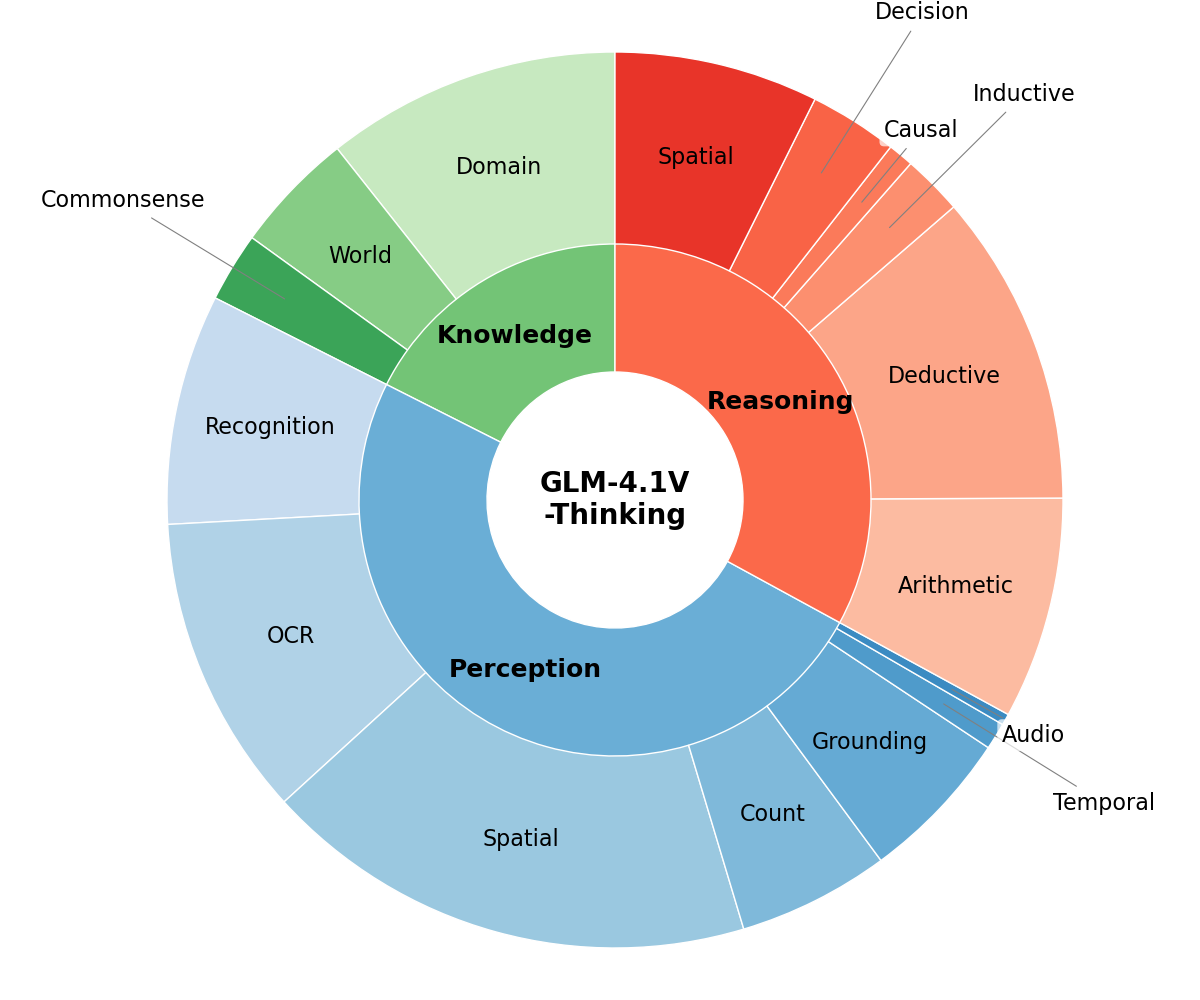}
  \end{subfigure}
  \vspace{-0.1in}
  \caption{Distribution of hallucination subcategories. Segment sizes correspond to cumulative scores.}
  \label{fig:hallu_pie}
\end{figure}

\begin{figure}[t]
  \centering
  \begin{subfigure}[b]{0.49\linewidth}
    \centering
    \includegraphics[width=\linewidth]{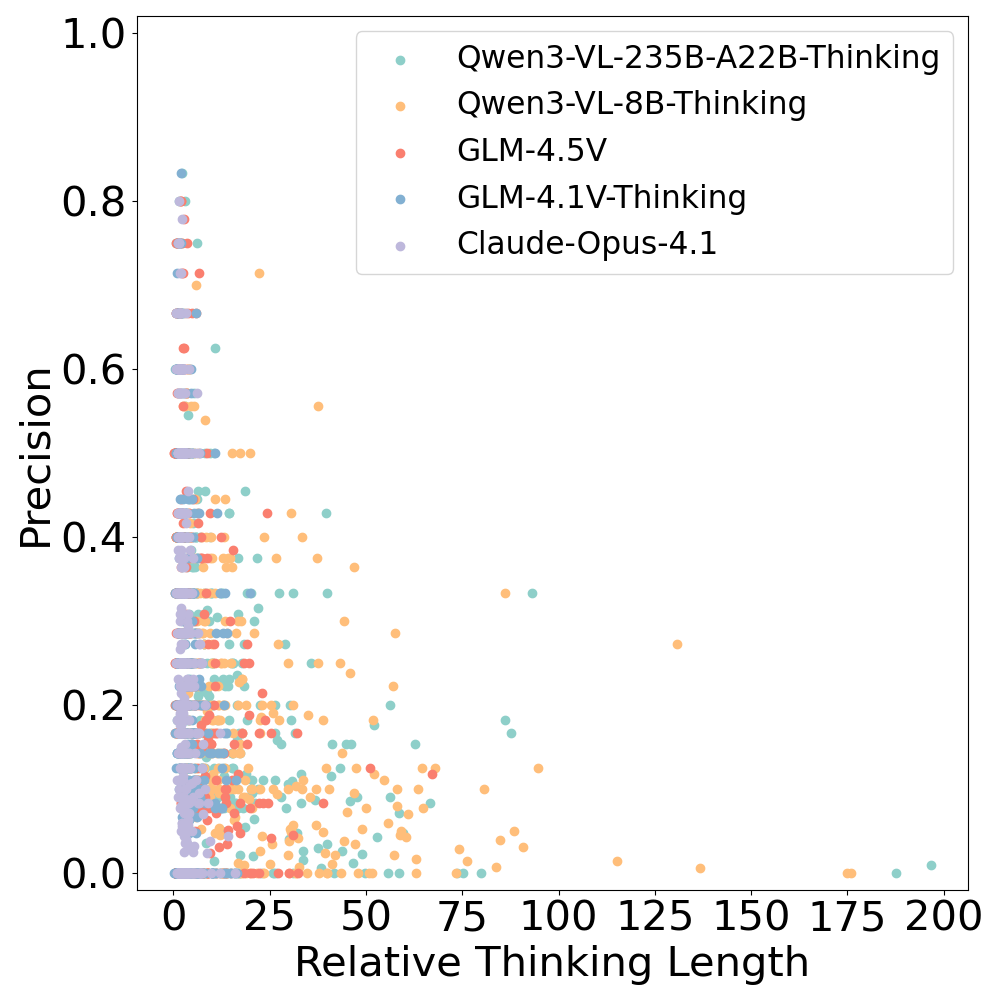}

  \end{subfigure}
  \hfill
  \begin{subfigure}[b]{0.49\linewidth}
    \centering
    \includegraphics[width=\linewidth]{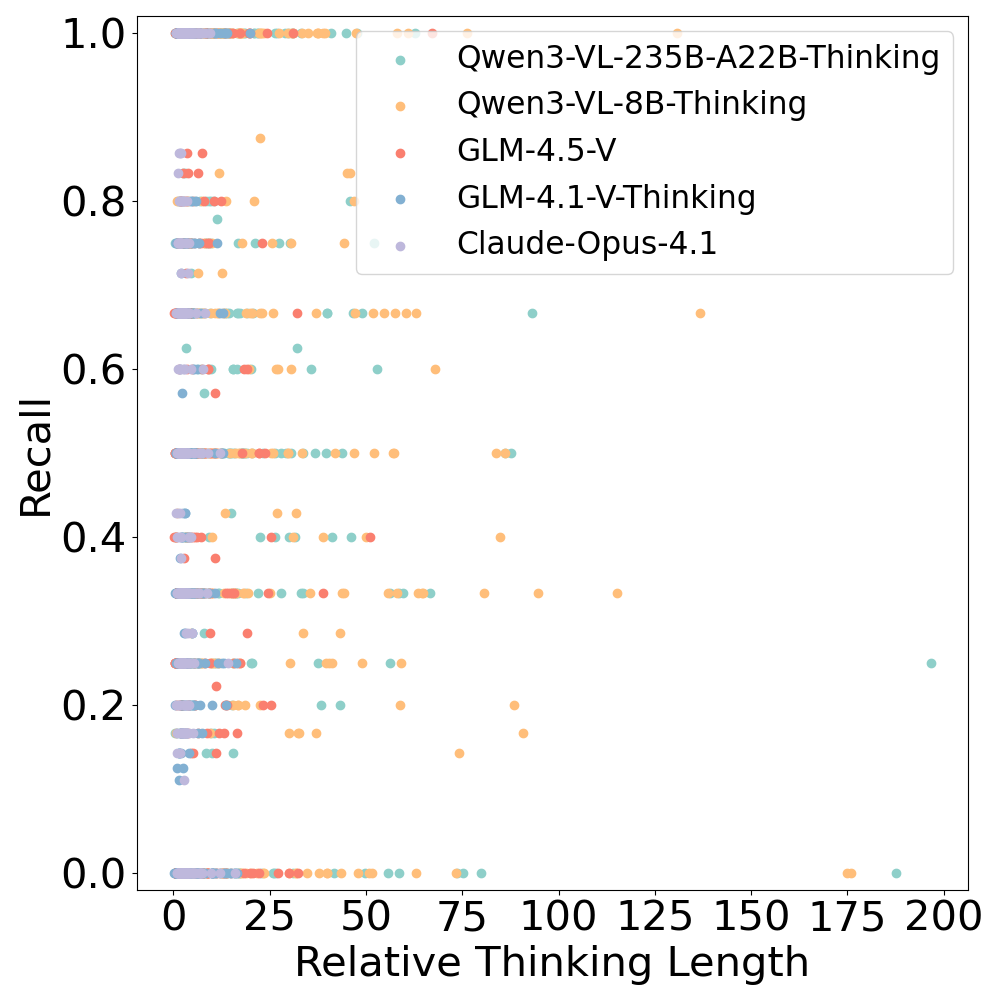}
  \end{subfigure}
  \vspace{-0.1in}
  \caption{Relationship between \textit{thinking length} and \textit{quality}, where the X-axis represents the relative length of thinking tokens, Y-axis represents step-level precision (\textbf{left}) and recall (\textbf{right}).}
  \label{fig:length_thinking_quality}
\end{figure}

\textbf{Analysis of Hallucination Subcategories.}
To investigate finer-grained hallucination behaviors, we further analyze the distribution of hallucination subcategories across major dimensions in the \bench taxonomy.
As illustrated in \cref{fig:hallu_pie}, we show the result for GLM-4.1V-thinking.
This model is chosen since we can evaluate its performance on video data,
making its analysis more representative of model behavior across the full dataset.

Several observations can be made:
(1) Hallucinations in the Perception dimension account for roughly half of the total hallucination score.
Within this dimension, \textbf{Spatial} hallucinations dominate, and a similarly high proportion of \textbf{Spatial} reasoning hallucinations occurs in the Reasoning dimension, highlighting deficiencies in spatial perception and reasoning of current reasoning MLLMs.
(2) In the Knowledge dimension, hallucinations related to \textbf{Domain Knowledge} are most prevalent.
(3) In the Reasoning dimension, besides spatial reasoning shortcomings, notable hallucinations occur in \textbf{Arithmetic} and \textbf{Deductive} subcategories.

In summary, our analysis indicates that although perception hallucinations are more common in multimodal reasoning, they are less detrimental to final answer correctness than reasoning hallucinations.
The distribution of hallucination subcategories is shaped by the dataset design and the model's inherent capabilities.
These findings validate the comprehensiveness and utility of our hallucination taxonomy and \bench for evaluating thinking hallucinations.

\subsection{CoT Efficiency}
\label{subsec:thinking_analysis}

While \cref{tab:overall_performance} highlights that most explicit thinking reasoning MLLMs achieve balanced Knowledge, Perception, and Reasoning capabilities, step-level metrics reveal a clear pattern: recall is higher than precision.
For instance, Qwen3-VL-235B attains $58.34\%$ recall but only $23.38\%$ precision.
Models generate more reasoning steps than necessary, covering most ground-truth steps but with many redundant or irrelevant ones.
This motivates a deeper analysis of the relationship between step-level length and effectiveness.

We choose five representative models evaluated above and plot their results on $500$ sampled data points. 
\cref{fig:length_thinking_quality} presents the relationship between the relative length and the thinking quality, measured by precision and recall.

\textbf{Precision-Length Scatter}:
As illustrated in \cref{fig:length_thinking_quality} (left), we observe a striking right-triangular distribution for step precision against the relative thinking length.
This pattern reveals a fundamental performance constraint: 
(1) High precision scores ($Recall>0.5$) are almost exclusively concentrated in the region of short relative thinking length ($Length>25$).
(2) The upper-right quadrant of the plot, which represents long yet highly precise thinking, is conspicuously sparse.
This triangular pattern reflects an inherent trade-off implied by the definition of precision.
As the number of generated reasoning steps increases, the denominator in precision grows faster than the numerator, since not all additional steps contribute to true positives.

\textbf{Recall-Length Scatter}:
\cref{fig:length_thinking_quality} (right) shows the relationship between step recall and thinking length.
Two patterns emerge:
(1) Points cluster along $Recall = 1.0$ or $Recall = 0.0$, indicating models either fully capture necessary steps or fail to align with them.
(2) An arc-like upper boundary marks the maximum achievable recall for each relative thinking length, with points distributed below it.
The arc's curvature varies across models, revealing trade-offs between thoroughness and conciseness.
High recall does not require excessive length, but longer chains may deviate from the correct path and miss key steps beyond a threshold.

\begin{table}[t]
\centering
\caption{Correlation between hallucination rates and relative token length, where $r_p$ represents Pearson, and $r_s$ represents Spearman. Significance: $^{*}p<0.05$, $^{**}p<0.01$, $^{***}p<0.001$.}
\label{tab:hallucination_length}
\resizebox{\linewidth}{!}{
\newcolumntype{d}[1]{D{.}{.}{#1}}
\begin{tabular}{l llll}
\toprule
\multirow{2}{*}{\textbf{Model}} & \multicolumn{2}{c}{\textbf{Perception}} & \multicolumn{2}{c}{\textbf{Reasoning}}\\
                    \cmidrule(r){2-3} \cmidrule(lr){4-5} 
                    & \multicolumn{1}{c}{\textbf{$r_p$}} & \multicolumn{1}{c}{\textbf{$r_s$}} & \multicolumn{1}{c}{\textbf{$r_p$}} & \multicolumn{1}{c}{\textbf{$r_s$}} \\
\midrule
GLM-4.1V-Thinking                   & $0.06^{*}$   & $0.10^{***}$ & $0.05^{*}$    & $0.09^{**}$    \\
GLM-4.5V                   & $0.07^{*}$   & $0.06^{*}$   & $0.14^{***}$  & $0.16^{***}$   \\
InternVL3.5-241B            & $0.10^{***}$ & $0.10^{***}$ & $0.23^{***}$  & $0.18^{***}$   \\
Doubao-seed-1.6             & $0.12^{***}$ & $0.13^{***}$ & $0.11^{***}$  & $0.13^{***}$   \\
Claude-Opus-4.1             & $0.05$       & $0.05$       & $0.09^{**}$   & $0.07^{**}$    \\
\bottomrule
\end{tabular}
}
\end{table}

\textbf{Hallucination-Length Correlation}:
To investigate the effect of reasoning length on hallucination in different dimensions, we compute both Pearson's and Spearman's correlations between relative token length and hallucination rate for Perception and Reasoning.
As shown in \cref{tab:hallucination_length}, all models exhibit a weak positive correlation with reasoning hallucinations ($r<0.25, p<0.01$), while perception correlations are generally non-significant ($r<0.12$).
This suggests that longer reasoning chains amplify reasoning hallucinations, with overthinking occurring after perception.

We observe that smaller models exhibit weaker correlations, suggesting more stable reasoning despite shorter intermediate CoTs.
Larger open-source models, such as InternVL3.5-241B-A28B, show stronger positive correlations, indicating longer chains are more prone to drift.
Closed-source models like Claude-Opus-4.1 better control hallucinations relative to chain length, reflecting effective mechanisms against overthinking.
Overall, hallucination tendencies depend on reasoning length, model scale, and post-training thinking preferences.
Additional results are in Appendix \ref{app:add_experiments}.

\section{Related Work}
\label{sec:related_work}

\textbf{Reasoning MLLMs.}
Recent reasoning LLMs, such as DeepSeek-R1~\cite{openai2024o1,deepseekai2025deepseekr1}, demonstrate that reinforcement learning effectively enhances multi-step reasoning capabilities. 
After exploring different approaches~\cite{chen2025sft,liu2025visualrft}, this post-training framework has been extended to MLLMs~\cite{yang2025r1onevision,yang2025r}.
However, recent studies suggest that the introduction of intermediate CoTs is not always beneficial to reasoning LLMs and MLLMs~\cite{su2025underthinking,tian2025thought}.
With the emergence of increasingly powerful reasoning MLLMs~\cite{vteam2025glm45v,qwen3technicalreport,wang2025internvl3_5}, it becomes essential to systematically evaluate both the quality of the thinking processes and the hallucinations produced within them.

\textbf{Mitigation of Multimodal Hallucination.}
As research on multimodal hallucination deepens, mitigating hallucinations has become an inevitable focus.
Training-free methods aim to reduce hallucinations without modifying model parameters~\cite{zhao2024mitigating, zheng2024thinking, tang2025seeing}.
Post-training offers a new potential solution for mitigating training-based hallucination, in addition to data augmentation and improved pretraining and fine-tuning methods~\cite{sarkar2024mitigating, li2025mitigating}.
Some researchers attempt to mitigate multimodal hallucination by improving the reinforcement learning method~\cite{wan2025srpo,2025arXiv250524238D,tian2025thought,fu-etal-2025-mitigating}.

\textbf{Multimodal Benchmark.}
Before MLLMs, early multimodal datasets such as COCO~\cite{lin2015coco}, VQAv2~\cite{goyal2017makingvvqamatter}, 
and GQA~\cite{hudson2019gqa} focus on the perception dimension, evaluating whether models correctly interpret visual information.
With the advent of MLLMs, recent works have introduced more diverse focuses:
Comprehensive benchmarks, such as MMBench~\cite{liu2024mmbench}, MM-Vet-v2~\cite{yu2024mmvetv2}, MMMU-pro~\cite{yue2023mmmu}, RBench-V~\cite{guo2025rbenchv}, and ZeroBench~\cite{roberts2025zerobench} assess overall multimodal reasoning.
Domain-specific benchmarks target specialized skills
such as mathematical reasoning~\cite{wang2024measuring,lu2024mathvista,wang2025mathcodervl},
spatial reasoning~\cite{chen2024spatialvlm,yang2025vsibench,jia2025omnispatial},
video understanding~\cite{fu2025videomme,hu2025videommmu, wang2025lvbench}, \textit{etc}.
Benchmarks for multimodal hallucination, 
including POPE~\cite{li2023pope} and HallusionBench~\cite{2023arXiv231014566G}, identify and analyze hallucinations in multimodal tasks. 
Post-training enables MLLMs to provide intermediate CoTs, 
and works such as MIRAGE~\cite{2025arXiv250524238D} and RH-Bench~\cite{2025arXiv250521523L} further analyze the hallucinations.
However, these studies lack a systematic taxonomy and have limited coverage.
This gap motivates \bench, which aims to establish a more comprehensive taxonomy and evaluation framework for multimodal hallucination during thinking.

\section{Conclusion}
\label{sec:conclusion}
We introduce \bench for evaluating multimodal hallucinations in the intermediate CoTs.
\bench provides a fine-grained hallucination taxonomy and offers an automated, reproducible evaluation framework for measuring hallucinations in intermediate CoTs.
Overall, \bench helps the community study and mitigate multimodal thinking hallucinations, and guides the development of more reliable and capable reasoning MLLMs.

\section*{Impact Statement}
\label{sec:impact}
This paper aims to advance the evaluation of hallucinations in the intermediate CoTs of reasoning MLLMs.
The proposed dataset is intended solely for evaluation and benchmarking purposes, not for training models.
The automated evaluation pipeline introduces a judge model, which may introduce bias if treated as definitive evaluators; this risk is mitigated through human meta-evaluation in \cref{subsec:expr_setup}.

\bibliography{main}
\bibliographystyle{icml2026}

\clearpage
\appendix

\section{Taxonomy Details}
\label{app:taxonomy}
\subsection{Top Dimensions}
As introduced in \cref{subsec:benchmark_design},
knowledge, perception, and reasoning are the three core dimensions of capabilities that reasoning MLLMs should possess.
In \bench, we construct the step-level and rubric-level evaluation framework based on these dimensions.

\textbf{Knowledge.} 
This dimension focuses on the processing that models retrieve and apply facts or knowledge obtained from pre-training and stored in internal parameters, 
rather than information explicitly provided in the question statement or the multimodal inputs. 
A step in knowledge denotes the cognitive processing from \textit{External fact/knowledge} to \textit{Thinking step}.
Hallucinations in this dimension arise when the model fabricates, misremembers, or incorrectly applies such parametric knowledge.

\textbf{Perception.} 
This dimension assesses the processing that models use to extract or interpret information from text and multimodal inputs, primarily visual content.
A step in perception denotes the cognitive processing from \textit{Text/vision inputs} to \textit{Thinking step}.
Perception hallucinations occur when the model fails to identify correct visual elements, hallucinates nonexistent visual content, or ignores available visual evidence.

\textbf{Reasoning.}
This dimension captures the validity and soundness of the transformations performed based on all extracted information and retrieved knowledge.
A step in reasoning denotes the cognitive processing from \textit{Intermediate result} to \textit{Thinking step / The final answer}.
Reasoning hallucinations occur when the model introduces unsupported intermediate steps, invalid deductions, or logically inconsistent conclusions.

Although these three top-level dimensions ensure comprehensive coverage of core capabilities, 
they are not sufficient to capture the full granularity required for detailed hallucination evaluations.
To more precisely characterize the types of hallucinations that occur in multimodal reasoning and to provide actionable insights for model improvement, 
we further refine each dimension into a set of fine-grained subcategories.
In the following subsections, we define each subcategory and provide an illustrative rubric that specifies 
(1) how to determine whether the model performs the corresponding capability correctly, and 
(2) how to identify hallucinations associated with that subcategory.

\subsection{Knowledge Subcategories}
\textbf{(K1) Commonsense}: 
Commonsense refers to everyday intuitive understanding shared by humans.
Rubric example: \textit{Does the model apply commonsense knowledge to provide a reasonable estimate for the width of an adult's torso (e.g., around 40-50 cm)?}

\textbf{(K2) World Knowledge}:
World knowledge refers to stable and widely accepted factual information about the real world. 
Rubric example: \textit{Does the model possess the world knowledge to connect specific types of books (e.g., board books, concept books, nursery rhymes) to corresponding child developmental stages?}

\textbf{(K3) Domain Knowledge}:
 Domain knowledge refers to the specialized or technical knowledge required by a specific domain or professional field. 
Rubric example: \textit{Does the model possess the domain knowledge to accurately describe the visual characteristics of Downy mildew (fuzzy, downy growth) and Powdery mildew (white, powdery coating)?}

\subsection{Perception Subcategories}
\textbf{(P1) Recognition}:
Recognition refers to the ability to identify the objects or entities present in visual input, including their classification, attributes, and high-level features.
Rubric example: \textit{Does the model correctly identify the fit line as a horizontal line?}

\textbf{(P2) OCR}:
Optical Character Recognition (OCR) is the ability to read and convert typed, handwritten, or printed text into textual tokens.
Rubric example: \textit{Does the model correctly read the title of the image as '5 STEPS TO WRITING AN SOP'?}

\textbf{(P3) Spatial}:
Spatial Perception is the ability to \textbf{directly} extract spatial information from the visual input, including absolute and relative positions, geometric layout, depth cues, et al.
Rubric example: \textit{Does the model accurately determine the spatial relationship, specifically that the yellow car is already in the destination lane and ahead of the driver's vehicle?}

\textbf{(P4) Count}:
Count is the ability to correctly perceive the number of target objects or events directly from the visual input.
Rubric example: \textit{Does the model accurately count the number of capybaras located in the area determined to be on the adult's left side?}

\textbf{(P5) Audio}:
Audio Perception refers specifically to the extraction of information from the soundtracks of video inputs in \bench, including speech (monologue and dialogue), background narration, and other non-speech sounds relevant to the task.
Rubric example: \textit{Does the model accurately transcribe or understand the player's spoken words, specifically 'I give up,' around timestamp 37:58?}

\textbf{(P6) Grounding}:
Grounding Perception refers to linking textual or symbolic references to their correct visual counterparts, including bounding-box, region referring, and resolving entity mentions.
Rubric example: \textit{Is the model able to locate the area marked with the number 11 on the Level 1 map?}

\textbf{(P7) Temporal}:
Temporal Perception refers to the ability to perceive and track dynamic changes over time in videos or image sequences.
Rubric example: \textit{Does the model correctly identify the start of the video segment explaining the situation in the 1990s, using either the audio cue at 08:00 or the on-screen title card at 08:05?}

\subsection{Reasoning Subcategories}
\textbf{(R1) Deductive}:
Deductive Reasoning refers to applying logical rules to derive conclusions from premises (previously obtained facts, perceptions, or intermediate reasoning results). 
Rubric example: \textit{Does the model correctly formulate the equation $p+q = 9$ based on the rule and the observed value for edge $PQ$?}

\textbf{(R2) Inductive}:
Inductive Reasoning refers to analogy, generalization, or probabilistic inference based on observed evidence.
Rubric example: \textit{Does the model infer from the visual evidence that the groups of people are separated based on their race?}

\textbf{(R3) Spatial}:
Spatial Reasoning refers to inferring or manipulating spatial relationships based on already perceived information.
Rubric example: \textit{Does the model successfully establish a new frame of reference from player 2's perspective, correctly identifying which direction is 'left' and which is 'right' for player 2?}

\textbf{(R4) Arithmetic}:
Arithmetic Reasoning refers to performing numerical operations based on extracted values or intermediate results.
Rubric example: \textit{Does the model correctly determine that there are 4 internal vertical sticks (6 total sticks - 2 end sticks) that contribute to the overlap?}

\textbf{(R5) Causal}:
Causal Reasoning refers to the identification of cause-and-effects relations, explains why an event happens, or predicts the possible outcomes.
Rubric example: \textit{Does the model correctly identify the causal relationship between the British Empire's large territorial holdings and the economic necessity of securing raw materials for its manufacturing sector (Option B)?}

\textbf{(R6) Decision}:
Decision-making Reasoning refers to the ability to select an option or conclusion based on preferences, constraints, or goals. 
Rubric example: \textit{Can the model differentiate between the primary economic driver of imperialism (acquiring raw materials, Option B) and a secondary benefit (creating markets for goods, Option C) to select the 'best' explanation?}

\textbf{(R7) Instructional}:
Instructional Reasoning refers to the ability to follow explicit instructions and execute the correct operations.
Rubric example: \textit{Does the model correctly parse the instruction 'click the UI element Home Office App' to understand that 'Home' is the target element?}

\section{Dataset Construction Details}
\label{app:construction}
The \cref{subsec:data_collection} in the main paper summarizes the high-level statistics of our benchmark.
This appendix provides a more detailed breakdown of data sources and construction steps that complement the statistics reported in the main text.

\begin{table}[t]
\centering
\caption{Statistics of data source in final \bench)}
\label{tab:data_source_stats}
\begin{tabular}{l rr}
\toprule
\textbf{Source Dataset} & \textbf{Count} & \textbf{Ratio} \\
\midrule
MathVision      & $288$     & $21.5\%$  \\
MM-vet-v2       & $150$     & $11.2\%$  \\
MMMU-pro        & $185$     & $13.8\%$  \\
HallusionBench  & $200$     & $14.9\%$  \\
Omni-Spatial    & $199$     & $14.9\%$  \\
CharXiv         & $100$     & $7.5\%$   \\
GUI-Agent       & $99$      & $7.4\%$   \\
Video-MME       & $119$     & $8.9\%$   \\
\midrule
Total           & $1340$    & $100.0\%$   \\
\bottomrule
\end{tabular}
\end{table}

\subsection{Details of Data Collection}
\textbf{Preprocessing of Image Input.}
For image data, we remove items whose task type is classified as \textit{caption generation}.
Such tasks typically emphasize the description of knowledge and perception, lacking the reasoning components required by \bench.

\textbf{Preprocessing of Video Input.}
For video data, we first ensure all the videos are still publicly accessible on YouTube at the time of download.
We then filter out very short clips (less than 2 minutes) to increase difficulty, as such clips often lack sufficient temporal content to support meaningful video understanding and reasoning.
To improve processing efficiency and maintain consistency in input scale, we further compressed the videos to a size within 20 MB without affecting recognizability.
Finally, we conducted two rounds of validation to confirm that the annotation model could not answer the question correctly without access to the corresponding video,
ensuring that the retained items require multimodal visual information rather than pure textual inference.

\subsection{Details of Annotation}

\textbf{Automatic Step Annotation.}
The annotation model, Gemini-2.5-pro, performs 2 annotation tasks in the construction of \bench:
(1) Primary step annotation and (2) Rubric annotation based on thinking steps.
We adopt API calls and maintain the unified setting for these two tasks.
Specifically, we activate the Thinking mode and keep the other parameters as the official default values.

\textbf{Human Step Annotation.}
Based on data with CoTs annotated by Gemini-2.5-pro, 
we construct an annotation platform and recruit human annotators to check the intermediate CoTs and ensure the correctness and necessity of each step.
In the annotation manual, the annotation procedure for each question is as follows:
(1) Sequentially review the annotated CoTs. 
Each step includes the step type, step description, step content, and step necessity, ensuring that all attributes are correctly annotated.
In addition, add any missing steps if needed.
(2) Evaluate the correctness of automated annotations, including the correctness of both the overall intermediate CoTs and the final answers.
(3) For questions that clearly exceed the annotators' capability or contain obvious ambiguity or errors, annotators are allowed to check the ``Too Complex'' option after spending \textbf{5 minutes} on the reasoning.
This stage of the human annotation process takes approximately 120 person-hours.
The compensation level for annotators is aligned with the average pay standards reported in comparable research annotation tasks.

\textbf{Automatic Rubric Annotation.}
Besides the same setting as automatic step annotation, we add three base rubric items for the annotation of each question:
\begin{itemize}
    \item{} Does the reasoning process use image or video information consistently and accurately? (Perception, 6 points)
    \item{} Does the reasoning process maintain logical coherence without gaps or contradictions? (Reasoning, 8 points)
    \item{} Is the reasoning process sufficient to support the final answer? (Reasoning, 9 points)
\end{itemize}
On the one hand, these rubric items correspond to core aspects of multimodal reasoning that \bench specifically aims to investigate.
On the other hand, these base rubric items serve as the scoring basis for score assignment.
For each question, these items are included in the rubric-level evaluation.

\textbf{Quality Controls of Step Annotation.}
To ensure the reliability of human annotations, the research team conducted a multi-stage quality control protocol after all annotators completed their tasks.
(1) Validation of ``Too Complex'' cases. 
We first inspect all samples labeled as \textit{Too Complex}.
If ambiguity or error is identified, we modify the question or options.
If a sample is confirmed to be genuinely beyond the scope of what \bench aims to evaluate, 
it is removed from the dataset.
(2) Annotator quality audit.
For each annotator, we randomly sampled 10\% of their annotated data for quality inspection.
We use the unqualified rate to measure quality, defined as the proportion of annotations that do not strictly follow the annotation guidelines.
Based on this rate, we adopted a three-tier policy:
\begin{itemize}
    \item{Unqualified rate $< 10\%$}. The research team directly corrects all mistakes found in the sampled data and accepts the remaining annotations as reliable.
    \item{$10\% \leq$ Unqualified rate $\leq 30\%$}. The annotator must revise all their annotations. After the revision, we perform another round of random sampling and quality auditing.
    \item{Unqualified rate $>30\%$}. We discard all data annotated by this annotator and relabel these data.
\end{itemize}

\textbf{Quality Controls of Rubric Annotation.}
After the completion of human annotations, Gemini-2.5-Pro is invoked again to generate rubrics for each data item.
A 10\% human sampling audit is conducted to ensure annotation quality, and we require the unqualified rate to remain below 30\%. 
Annotations exceeding this threshold are revised and rechecked accordingly.

\subsection{Details of Data Statistic}
\textbf{Source Distributions}
The final \bench contains 1340 samples; the detailed counts and ratios of data sources are shown in \cref{tab:data_source_stats}.

\begin{figure}[t]
  \centering
  \begin{subfigure}[b]{0.95\linewidth}
    \centering
    \includegraphics[width=\linewidth]{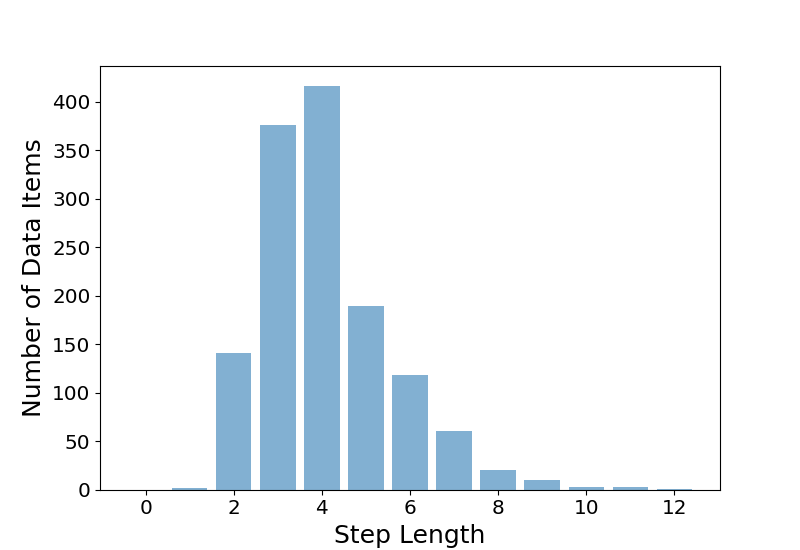}
  \end{subfigure}
  \caption{Distribution of the data items across different total thinking step counts.
  The X-axis denotes the number of thinking steps, and the Y-axis indicates the count of data items annotated with the corresponding number of steps.}
\label{fig:dist_step_cnt}
\end{figure}

\textbf{Step-level Statistic}.
Besides the step-level statistic shown in the main paper, we illustrate the distribution of data items across thinking step counts in \cref{fig:dist_step_cnt}.
As shown in the figure, the mode of total thinking step length is 4, with the maximum reaching 12 steps.
The modes align closely with the mean of 4.17, indicating a relatively balanced difficulty distribution across the data in \bench.

\begin{table}[t]
\centering
\caption{Rubric-level statistics across the three cognitive dimensions.}
\label{tab:rubric_level_stats}
\begin{tabular}{c r r}
\toprule
\textbf{Type}       &\textbf{Items / Q.}   &  \textbf{Scores / Q. }\\
\midrule
Knowledge           & $0.73$  & $4.42$  \\
Perception          & $2.51$  & $9.35$  \\
Reasoning           & $3.01$  & $15.63$ \\
\midrule
All Items           & $6.25$  & $29.40$ \\
\bottomrule
\end{tabular}
\end{table}

\begin{table*}[t]
\centering
\small
\caption{Accuracy of reasoning MLLMs across subsets. 
The names of source dataset are used to indicate the corresponding subsets. 
Column names correspond to the source datasets: 
\textbf{Math} (MathVision), 
\textbf{MM-vet} (MM-vet-v2), 
\textbf{MMMU} (MMMU-pro), 
\textbf{Hallusion} (HallusionBench), 
\textbf{Spatial} (Omni-Spatial), 
\textbf{GUI} (GUI-Agent), 
\textbf{Video} (Video-MME).
}
\label{tab:subset_acc}
\scalebox{0.95}{
\setlength{\tabcolsep}{7.5pt}
\begin{tabular}{lccccccrc}
\toprule
\multirow{2}{*}{\makebox[100pt][c]{\textbf{Model}}} & \multicolumn{8}{c}{\textbf{Subsets}} \\
                     \cmidrule(r){2-9} 
                     {} & \textbf{Math} & \textbf{MM-vet} & \textbf{MMMU} & \textbf{Hallusion} & \textbf{Spatial} & \textbf{CharXiv} & \textbf{GUI} & \textbf{Video}\\
\midrule     
                    MiniCPM-4.5V          & $42.71$ & $67.67$ & $63.04$ & $68.00$ & $50.75$ & $45.50$ & $0.00$ & $-$ \\
                    GLM-4.1V-Thinking     & $46.88$ & $68.00$ & $59.78$ & $71.50$ & $48.74$ & $54.00$ & $3.03$ & $42.02$ \\
                    Qwen3-VL-8B-Thinking  & $64.24$ & $70.67$ & $73.91$ & $69.00$ & $49.25$ & $54.00$ & $30.30$ & $-$ \\
                    InternVL3.5-8B        & $54.51$ & $69.67$ & $65.76$ & $65.00$ & $53.27$ & $46.00$ & $4.04$  & $-$ \\
                    
                    GLM-4.5V              & $65.97$ & $72.00$ & $72.28$ & $69.50$ & $51.76$ & $61.00$ & $43.43$ & $52.94 $ \\
                    Qwen3-VL-235B-A22B-Thinking & $\mathbf{76.39}$ & $\mathbf{75.67}$ & $\mathbf{85.33}$ & $72.00$ & $\mathbf{56.78}$ & $58.00$ & $\mathbf{65.66}$ & $-$ \\
                    InternVL3.5-241B-A28B & $64.24$ & $72.33$ & $75.00$ & $68.50$ & $53.77$ & $58.00$ & $34.34$ & $-$ \\
                    
                    Doubao-seed-1.6       & $65.28$ & $74.67$ & $79.35$ & $\mathbf{79.00}$ & $50.25$ & $59.00$ & $12.12$ & $\mathbf{61.34}$\\
                    Claude-Opus-4.1       & $63.02$ & $73.33$ & $78.80$ & $73.00$ & $47.74$ & $63.50$ & $13.13$ & $-$\\
                    Claude-Sonnet-4.5     & $62.85$ & $67.00$ & $79.89$ & $74.50$ & $51.26$ & $\mathbf{65.00}$ & $14.14$ & $-$\\
\midrule 
\rowcolor{gray!20}
                    o3                    & $71.53$ & $84.78$ & $77.67$ & $73.00$ & $58.59$ & $65.50$ & $4.04$ & $-$ \\
\rowcolor{gray!20}
                    GPT-5                 & $78.47$ & $74.67$ & $89.13$ & $75.50$ & $57.79$ & $65.00$ & $1.01$ & $-$ \\      
\rowcolor{gray!20}
                    Gemini-2.5-Flash      & $70.49$ & $80.67$ & $81.52$ & $72.00$ & $58.79$ & $67.50$ & $5.05$ & $71.42$ \\
\rowcolor{gray!20}
                    Gemini-2.5-Pro        & $79.51$ & $80.33$ & $85.33$ & $70.00$ & $55.78$ & $66.50$ & $7.07$ & $83.19$ \\
\bottomrule
\end{tabular}
}
\end{table*}

\textbf{Rubric-level Statistic}.
The statistics in \cref{tab:rubric_level_stats} summarize the full dataset at the rubric level. 
Note that the three base rubric items are not included in this statistic.
While we aim to maintain a relatively balanced coverage across s the three dimensions, the average number of items and scores per question is naturally higher for the perception and reasoning dimensions than Knowledge.
This distribution aligns with the focus of \bench on multimodal thinking and hallucinations.

\section{Automatic Evaluation Framework}
\label{app:framework}
As introduced in the main paper, 
\bench evaluates reasoning MLLMs at three levels using LLM-as-judge.
This section provides implementation details for reproducibility.

\subsection{Judge Model Configuration}
We use Qwen3-32B~\cite{qwen3technicalreport} as our judge model for evaluation, keeping the thinking mode enabled throughout the process.
The consistent inference and deployment settings are shown in \cref{tab:judge_parameters}.

\begin{table}[t]
\centering
\caption{Inference and deployment parameters for the judge model in the evaluation of \bench.}
\label{tab:judge_parameters}
\begin{tabular}{l l c}
\toprule
\textbf{Category}               &\textbf{Parameter}   &  \textbf{Value}\\
\midrule
\multirow{3}{*}{Inference}       & temperature  & $0.6$  \\
                                & top-p        & $0.95$  \\
                                & top-k        & $20$ \\
\midrule
\multirow{3}{*}{Deployment}     & max-token-length      & $32768$ \\
                                & cuda-graph-max-bs     & $128$   \\
                                & tensor-parallel-size  & $2$     \\

\bottomrule
\end{tabular}
\end{table}

\subsection{Judge Prompts}
\textbf{Answer Extraction and Judgment.}
Considering that the evaluated models may not strictly follow our instructions to produce structured final answers, we combine answer extraction and answer judgment into a single step.
The corresponding prompts for open-ended and multiple-choice questions are shown in \cref{fig:prompt_answer_oe} and \cref{fig:prompt_answer_mc}.

\textbf{Step Segmentation.}
We next describe the prompts used to segment the thinking steps.
Since intermediate thinking in reasoning MLLMs follows a fixed pattern learned from post-training, they cannot be reliably prompted to output step-by-step reasoning directly.
Therefore, we employ the judge model to segment the complete intermediate thinking process. 
The prompt is shown in \cref{fig:prompt_step_segment}.
For particularly long thinking processes that exceed the context window of the judge model, we first perform a preliminary split based on syntactic cues (e.g., keywords such as ``Wait''), then apply the segmentation procedure to each segment separately before concatenating the results.

\textbf{Step Match.}
Then, we employ the judge model to evaluate alignment between the ground-truth steps and the predicted steps, and the prompt is shown in \cref{fig:prompt_step_match}.
To improve efficiency and avoid completely context-free one-to-one comparisons, each evaluation provides the judge model with \textbf{one ground-truth step} and \textbf{all predicted steps}.
Notably, to prevent exceeding the context window, excessively long intermediate CoTs are first split into groups by step, and only a group with several steps is provided to the judge model at a time.
After all step groups have been evaluated, the results are merged to obtain the final results.

\textbf{Score Based on Rubric.}
Finally, we require the judge model to perform fine-grained scoring of the predicted intermediate CoTs based on an annotated rubric for each question.
Specifically, for each rubric item, the judge model first determines whether the item is satisfied. If not, the judge further evaluates whether the failure is attributed to hallucinations.
The scoring prompt is illustrated in \cref{fig:prompt_score}

\section{Significance and Reliability Analysis}
\label{app:significance}
To verify whether the qualitative trend in \cref{fig:acc_hallu} is statistically reliable, we further conduct hypothesis testing on the relationship between the correctness of final answers and hallucination categories.

We construct a $2\times2$ contingency table using counts of perception and reasoning hallucinations versus final correctness and apply both Fisher's exact test and the $\chi^2 $ independence test.
The results are shown in \cref{tab:hypo_test}.

Among the evaluated models, most exhibit statistically significant associations (e.g., $p<0.05$), indicating that reasoning hallucinations are more likely to lead to incorrect final answers. 
This trend aligns with our observations in \cref{fig:acc_hallu}.
However, for models without significant results, the reasoning hallucination error rate is relatively low, narrowing the gap between perception and reasoning hallucinations.
This small difference reduces the effect size, leading to weak statistical significance despite the same qualitative trend.

\begin{table}[t]
\centering
\caption{Results of hypothesis testing on the relationship between the final correctness and hallucination categories. 
Significance: $^{*}p<0.05$, $^{**}p<0.01$, $^{***}p<0.001$.
$\mathbf{OR}$ represents the Fisher odds ratio.
$\mathbf{Err_P}$ and $\mathbf{Err_R}$ denote the empirical error rates (\%) caused by perception and reasoning hallucinations.
Qwen3-VL-235B denotes Qwen3-VL-235B-A22B-Thinking.}
\label{tab:hypo_test}
\resizebox{\linewidth}{!}{
\newcolumntype{d}[1]{D{.}{.}{#1}}
\begin{tabular}{l llll}
\toprule
\textbf{Category}               &$\mathbf{\chi ^2}$   &  $\mathbf{OR}$       & $\mathbf{Err_P}$   & $\mathbf{Err_R} $       \\
\midrule
GLM-4.1V-Thinking               &  $0.1$              & $1.2$                &     $60.8$         &         $65.1$          \\
Qwen3-VL-8B-Thinking            &  $5.5^{*}$          & $2.8^{*}$            &     $60.7$         &        $81.3$           \\
InternVL3.5-8B                  &  $3.7$              & $3.7$                &     $54.6$         &        $73.8$           \\
GLM-4.5V                        &  $4.8^{*}$          & $3.1^{*}$            &     $55.6$         &        $79.4$           \\ 
Qwen3-VL-235B                   &  $10.3^{**}$        & $4.3^{**}$           &     $48.5$         &        $80.0$           \\
InternVL3.5-241B-A28B           &  $4.3^{*}$          & $2.7^{*}$            &     $64.9$         &        $83.3$           \\
Doubao-seed-1.6                 &  $0.6$              & $1.5$                &     $59.1$         &        $68.8$           \\ 
Claude-Opus-4.1                 &  $9.5^{**}$         & $3.9^{**}$           &     $59.4$         &        $85.2$           \\

\bottomrule
\end{tabular}
}
\end{table}

\section{Additional Experimental Results}
\label{app:add_experiments}
The accuracy for each model across different subsets is presented in \cref{tab:subset_acc}.

\textbf{Analysis on GUI-Agent.}
Among all the subsets, \textbf{GUI-Agent} exhibits the largest performance gap across models. 
The best-performing model achieves an accuracy of 65.66\%, while the weakest model obtains 0\%.
Even some models that perform strongly on most other tasks show very low accuracy on this subset, such as Gemini-2.5-pro.

This task requires the model to identify the target element in the visual input and output its position in the format \textit{[y1, x1, y2, x2]}.
In the accuracy calculation, we verify outputs using two matching strategies (absolute coordinates or normalized relative coordinates), and a prediction is considered correct if either format matches.

However, we observe two typical failure patterns:
(1) Format invalidity: Some models fail to output a valid coordinate tuple and instead reproduce the literal string ``[y1, x1, y2, x2]'', indicating that they have not learned this structured output format.
(2) Large spatial deviation: Even when the format is correct, the predicted bounding box often differs substantially from the ground truth, resulting in incorrect predictions.

We believe this is due to a lack of training data or alignment objectives for this grounding-based coordinate prediction task in many existing foundation models.
This limitation may limit the applicability of such models in downstream GUI-agent scenarios that require precise spatial localization.

\section{Limitations and Future Work}
\label{app:limitations}
This work has the following main limitations:

\textbf{Dependence of Judge Model.}
The proposed evaluation framework of \bench relies heavily on a single LLM as the automatic judge.
Although we design structured prompts and clear evaluation rules, the judge model may still introduce biases, including inherent preferences, reasoning errors, and hallucinations.
Moreover, the judge evaluates perception and reasoning steps based solely on intermediate CoTs without direct access to the original visual inputs, which may limit fidelity to the actual multimodal evidence.
In future work, we plan to explore a multi-judge setup that combines complementary models.
For example, an LLM for reasoning-focused evaluation and two MLLMs for perception-sensitive evaluation. 
Such a voting mechanism could improve robustness and reduce bias or hallucinations introduced by relying on a single judge.

\textbf{Multimodal Task Coverage.}
While \bench includes a wide variety of datasets, the majority of evaluated tasks still follow the question-answering paradigm.
Emerging multimodal reasoning challenges, such as interactive GUI navigation, agent-based tool calling, and embodied perception, are not systematically covered.
Although \bench includes GUI grounding tasks, it does not fully evaluate multimodal reasoning and hallucinations in interactive or simulated environments.
Future work could focus on interactive, tool-augmented, and embodied settings, which may also yield novel subcategories of hallucinations that are not captured by current QA-centric multimodal tasks.

\begin{figure*}[htbp]
\begin{tcolorbox}[
  colframe=cyan!75!black, 
  colback=white, 
  coltitle=white, 
  colbacktitle=cyan!75!black, 
  width=\linewidth, 
  arc=2mm, 
  auto outer arc, 
  boxrule=0.5pt, 
  left=10pt, 
  right=10pt, 
  top=5pt, 
  bottom=5pt, 
  title=\textbf{Prompt for answer extraction and judgment for open-ended questions}, 
  fonttitle=\bfseries,
]
\small
You are an expert judge for answer correctness.\\
You will be provided: (1) A multimodal question; (2) The ground truth answer; (3) A predicted
response given by a model.\\
\textbf{Your task}:\\
1. If the model's reponse includes reasoning or multiple steps, extract the final answer from the response, don't change the wording of the answer, just extract it; \\
2. Compare the model's final answer with the ground truth answer; \\
3. Only return you result **strictly in JSON format**. Do not output any other information.\\
\textbf{Rules}:\\
1. Ignore any reasoning or irrelevant information in the response;\\
2. Consider two answers semantically equivalent if they represent the same value or entity, even if formatting, unit, or minor wording differs, it is considered as MATCH (e.g., `twelve' and `12', `alpha' and `$\alpha$');\\
3. If the predicted answer contains some, but not all, elements of the ground truth answer, or includes extra incorrect elements along with correct ones, mark it as PARTIAL;\\
4. Be strict about the meaning - if the predicted answer is completely not semantically equivalent to the ground truth answer, it is considered as MISMATCH;\\
5. If no final answer is provided in the response, it is considered as NO ANSWER.\\
\textbf{Examples}: \\
\verb!|! Question \verb!|! Ground truth answer \verb!|! Prediction \verb!|! Answer Match \verb!|! \\
\verb!|! ---------- \verb!|! -------------------- \verb!|! ------------ \verb!|! -------------- \verb!|! \\
\verb!|! What is the value of x in the equation? \textless image\textgreater \verb!|! 0 \textless AND\textgreater 1 \verb!|! x=2 \verb!|! MISMATCH \verb!|! \\
\verb!|! What is the value of x in the equation? \textless image\textgreater \verb!|! 0 \textless AND\textgreater 1 \verb!|! x=0 or 2 \verb!|! PARTIAL \verb!|! \\
\verb!|! What is the value of x in the equation? \textless image\textgreater \verb!|! 0 \textless AND\textgreater 1 \verb!|! x=0 or 1 or 2 \verb!|! PARTIAL \verb!|! \\
\verb!|! What is the value of x in the equation? \textless image\textgreater \verb!|! 0 \textless AND\textgreater 1 \verb!|! x=1 or 0 \verb!|! MATCH \verb!|! \\
\verb!|! Which curve is the highest at x = 0? \textless image\textgreater \verb!|! $\beta_1$ \verb!|! $\beta _2$ \verb!|! MISMATCH \verb!|! \\
\verb!|! Which curve is the highest at x = 0? \textless image\textgreater \verb!|! $\beta_1$ \verb!|! Curve $\beta_1$ \verb!|! MATCH \verb!|! \\
\verb!|! Which curve is the highest at x = 0? \textless image\textgreater \verb!|! $\beta_1$ \verb!|! $beta_1$\verb!|! MATCH \verb!|! \\
\verb!|! What is the length(cm) of Edge AC? \textless image\textgreater \verb!|! 10 cm \verb!|! 10.00 \verb!|! MATCH \verb!|! \\
\verb!|! What is the length(cm) of Edge AC? \textless image\textgreater \verb!|! 10 cm \verb!|! AC=10 cm \verb!|! MATCH \verb!|! \\
\textbf{JSON schema}: \{ \\
\hspace*{2em}    ``pred\_answer'': str, \\
\hspace*{2em}   ``answer\_match'': str,  \# ``MATCH'', ``PARTIAL'' or ``NO ANSWER'';\\
\} \\\
\textbf{Question}: \textcolor{violet}{[Q]} \\
\textbf{Ground truth answer}: \textcolor{violet}{[GT]} \\
\textbf{Predicted response}: \textcolor{violet}{[PRED]} \\
\end{tcolorbox}

\caption{Prompt for answer extraction and judgment for open-ended questions.}
\label{fig:prompt_answer_oe}
\end{figure*}

\begin{figure*}[htbp]
\begin{tcolorbox}[
  colframe=cyan!75!black, 
  colback=white, 
  coltitle=white, 
  colbacktitle=cyan!75!black, 
  width=\linewidth, 
  arc=2mm, 
  auto outer arc, 
  boxrule=0.5pt, 
  left=10pt, 
  right=10pt, 
  top=5pt, 
  bottom=15pt, 
  title=\textbf{Prompt for answer extraction and judgment for multi-choice questions}, 
  fonttitle=\bfseries, 
]
\small
You are an expert judge for answer correctness.\\
You will be provided: (1) A multimodal question; (2) Options of the question; (3) The ground truth answer; (4) A predicted response given by a model; \\
\textbf{Your task}:\\
1. If the model's reponse includes reasoning or multiple steps, extract the final answer from the response, only keep the uppercase letter of the chosen option (A, B, C, ...); \\
2. Compare the model's final answer with the ground truth answer;\\
3. Only return you result **strictly in JSON format**. Do not output any other information.\\
\textbf{Rules}: \\
1. Ignore any reasoning or irrelevant information in the response;\\
2. Consider two answers semantically equivalent if they represent the same value or entity, even if formatting, unit, or minor wording differs, it is considered as MATCH (e.g., `twelve' and `12', `alpha' and `$\alpha$');\\
3. Be strict about the meaning - if the predicted answer is completely not semantically equivalent to the ground truth answer, it is considered as MISMATCH;\\
4. If no final answer is provided in the response, it is considered as NO ANSWER;\\
\textbf{Examples}: \\
\verb!|! Question \verb!|! Options \verb!|! Ground truth \verb!|! Prediction \verb!|! Answer Match \verb!|! \\
\verb!|!----------\verb!|!---------\verb!|!--------------\verb!|!------------\verb!|!--------------\verb!|! \\
\verb!|! Where is the photo taken? \textless image\textgreater \verb!|! A.London, B.Paris, ... \verb!|! A \verb!|! B \verb!|! MISMATCH \verb!|! \\
\verb!|! Where is the photo taken? \textless image\textgreater \verb!|! A.London, B.Paris, ... \verb!|! A \verb!|! London \verb!|! MATCH \verb!|! \\
\verb!|! Where is the photo taken? \textless image\textgreater \verb!|! A.London, B.Paris, ... \verb!|! A \verb!|! A. London \verb!|! MATCH \verb!|! \\
\verb!|! Where is the photo taken? \textless image\textgreater \verb!|! A.London, B.Paris, ... \verb!|! A \verb!|! A \verb!|! MATCH \verb!|! \\
\textbf{JSON schema}: \{ \\
\hspace*{2em}    ``pred\_answer'': str, \\
\hspace*{2em}   ``answer\_match'': str,  \# ``MATCH'', ``PARTIAL'' or ``NO ANSWER'';\\
\} \\\
\textbf{Question}: \textcolor{violet}{[Q]} \\
\textbf{Options}: \textcolor{violet}{[OPT]} \\
\textbf{Ground truth answer}: \textcolor{violet}{[GT]} \\
\textbf{Predicted response}: \textcolor{violet}{[PRED]} \\
\end{tcolorbox}
\caption{Prompt for answer extraction and judgment for multi-choice questions.}
\label{fig:prompt_answer_mc}
\end{figure*}

\begin{figure*}[htbp]
\begin{tcolorbox}[
  colframe=cyan!75!black, 
  colback=white, 
  coltitle=white, 
  colbacktitle=cyan!75!black, 
  width=\linewidth, 
  arc=2mm,
  auto outer arc,
  boxrule=0.5pt,
  left=10pt, 
  right=10pt,
  top=5pt,
  bottom=15pt,  
  title=\textbf{Prompt for step segmentation}, 
  fonttitle=\bfseries,
]
\small
You are a step segmentation assistant.\\
Your task is to segment the input thinking text into a list of steps.\\
\textbf{Rules}: \\
1. Keep the original wording of the input text. Do not rewrite or summarize the content. \\
2. Split when the reasoning begins a new action, inference, perception, or reflection. \\
3. Include mistakes and self-corrections, do not delete. \\
4. Merge sentences or clauses that continue the same line of thought into a single step. \\
5. Do not create a new step for phrases that express only hesitation, emotion, or self-reflection, such as ``Let me think.'', ``I'm stuck'', etc. Keep them together with the following thinking step. \\
6. Typically expect 3–10 steps.\\
7. Return only valid JSON, no explanations. Do not add any text outside the JSON. \\
\textbf{JSON schema}: [ \\
\hspace*{2em}   \{ \\
\hspace*{4em}       ``step\_id'': int, \\
\hspace*{4em}      ``step\_content'': str, \\
\hspace*{2em}    \}, \\
] \\
\textbf{Example}:\\
Input: `` Got it, let's analyze it. First, let's look at the image, the image shows a clover leaf with yellow spots. Then, let's think about the possible reasons for the yellow spots. I think the yellow spots are caused by viral infections, because the leaves are yellow. Hmm, I'm not sure. Wait, the yellow spots are not caused by viral infections, but by nutrient deficiency. So the final answer is B. nutrient deficiency.'' \\
Output: [ \\
\hspace*{2em}    \{ \\
\hspace*{4em}       ``step\_id'': 1, \\
\hspace*{4em}       ``step\_content'': ``Got it, let's analyze it. First, let's look at the image, the image shows a clover leaf with yellow spots.'' \\
\hspace*{2em}    \},\\
\hspace*{2em}    \{ \\
\hspace*{4em}        ``step\_id'': 2, \\
\hspace*{4em}       ``step\_content'': ``Then, let's think about the possible reasons for the yellow spots. I think the yellow spots are caused by viral infections, because the leaves are yellow.'' \\
\hspace*{2em}    \}, \\
\hspace*{2em}    \{ \\
\hspace*{4em}       ``step\_id'': 3, \\
\hspace*{4em}       ``step\_content'': ``Hmm, I'm not sure. Wait, the yellow spots are not caused by viral infections, but by nutrient deficiency.''\\
\hspace*{2em}    \},\\
\hspace*{2em}    \{ \\
\hspace*{4em}        ``step\_id'': 4, \\
\hspace*{4em}        ``step\_content'': ``So the final answer is B. nutrient deficiency.'' \\
\hspace*{2em}    \} \\
] \\
\textbf{Raw thinking text}: \textcolor{violet}{[T]}
\end{tcolorbox}
\caption{Prompt for step segmentation}
\label{fig:prompt_step_segment}
\end{figure*}

\begin{figure*}[htbp]
\begin{tcolorbox}[
  colframe=cyan!75!black, 
  colback=white,
  coltitle=white, 
  colbacktitle=cyan!75!black, 
  width=\linewidth,
  arc=2mm, 
  auto outer arc, 
  boxrule=0.5pt, 
  left=10pt,
  right=10pt, 
  top=5pt, 
  bottom=15pt, 
  title=\textbf{Prompt for step match}, 
  fonttitle=\bfseries,
]
\small
You are an expert reasoning aligment judge. \\
You will be provided: (1) A multimodal question; (2) The ground truth answer; (3) One ground truth reasoning step; (4) A list of reasoning steps produced by a model; \\
\textbf{Your task}: \\
1. Compare the ground truth step with each of the predicted steps one by one, consider both semantic content and cognitive function;\\
2. Identify whether one of the predicted steps expresses the *same or similar meaning* as the ground truth step, if so, return the step\_id of the most similar predicted step;\\
\textbf{Matching Rules}:\\
1. **MATCH**: The predicted step expresses the same intent and type (knowledge, perception, reasoning) as the ground truth steps, conveys equivalent meaning or intermediate results, and is factually correct or logically consistent, even if formatting or minor wording differs (e.g., ``twelve'' vs ``12'', ``alpha'' vs ``$\alpha$''). \\
2. **PARTIAL**: The predicted step partially covers the same meaning or cognitive type, but misses key details,contains minor reasoning errors, or omits important intermediate results. \\
3. **MISMATCH**: None of the predicted steps express similar meaning or serve the same function as the ground truth step, or all are completely wrong.\\
\textbf{Output Rules}: \\
1. For ``MATCH'' or ``PARTIAL'', return the ID of the most similar predicted steps. \\
2. For ``MISMATCH'', return -1; \\
3. Only return you result **strictly** in JSON format. Do not output any other information. \\
\textbf{JSON schema}: \{ \\
\hspace*{2em}    ``best\_match\_step\_id'': int,    \# -1 if MISMATCH \\
\hspace*{2em}    ``step\_match'': str,     \# ``MATCH'', ``PARTIAL'', or ``MISMATCH''\\
\}\\
\textbf{Question}: \textcolor{violet}{[Q]} \\
\textbf{Ground truth answer}: \textcolor{violet}{[GT]} \\
\textbf{Ground truth step}: \textcolor{violet}{[GT\_STEP]} \\
\textbf{Predicted steps}: \textcolor{violet}{[PRED\_STEPS]} \\
\end{tcolorbox}
\caption{Prompt for step match.}
\label{fig:prompt_step_match}
\end{figure*}

\begin{figure*}[htbp]
\begin{tcolorbox}[
  colframe=cyan!75!black, 
  colback=white, 
  coltitle=white, 
  colbacktitle=cyan!75!black,
  width=\linewidth, 
  arc=2mm,
  auto outer arc,
  boxrule=0.5pt, 
  left=10pt, 
  right=10pt, 
  top=5pt,
  bottom=15pt, 
  title=\textbf{Prompt for scoring based on rubric},
  fonttitle=\bfseries, 
]
\small
You are a expert judge for multimodal reasoning quality. \\
You will be provided: (1) A multimodal question; (2) The ground truth answer; (3) The ground truth reasoning steps;
(4) A list of reasoning steps given by a model;
(5) A rubric. Each item specifies what the model should achieve, and link to one or more reasoning steps. \\
\textbf{Your task}:\\
1. For each item, determine whether the model's reasoning steps fully and correctly satisfies the question; \\
2. Answer only ``YES'' if the rubric item is satisfied, otherwise answer ``No'';\\
3. If the judgment is ``No'', and the reason is that the model produced hallucination, set the ``hallucination'' to true, otherwise set it to false;\\
4. List the IDs of the reasoning steps you used as evidence for your decision;\\

\textbf{Rules}: \\
1. The order of the judgments in response should be the same as the order of the criteria in the rubric;\\
2. Only return you result **strictly in JSON format. Do not output any other information.\\
\textbf{JSON schema}: [ \\
\hspace*{2em}     \{ \\
\hspace*{4em}         ``judgment'': str, \# ``YES'' or ``No''; \\
\hspace*{4em}         ``hallucination'': bool,  \# true if hallucination detected (judgment is ``NO''), false otherwise; \\
\hspace*{4em}         ``evidence\_step\_ids'': list[int], \# The IDs of the reasoning steps used as evidence; \\
\hspace*{2em}     \}, \\
] \\
\textbf{Question}: \textcolor{violet}{[Q]} \\
\textbf{Ground truth answer}: \textcolor{violet}{[GT]} \\
\textbf{Ground truth steps}: \textcolor{violet}{[GT\_STEPS]} \\
\textbf{Reasoning steps}: \textcolor{violet}{[PRED\_STEPS]} \\
\textbf{Rubric}: \textcolor{violet}{[R]} \\

\end{tcolorbox}
\caption{Prompt for scoring based on rubric.}
\label{fig:prompt_score}
\end{figure*}

\end{document}